\documentclass[11pt]{article}

\usepackage[T1]{fontenc}
\usepackage{charter}
\usepackage[scaled=0.92]{helvet}

\usepackage[a4paper,margin=2cm]{geometry}
\usepackage[table]{xcolor}
\definecolor{ratinghead}{HTML}{1B5E20}
\definecolor{ratingdark}{HTML}{C8E6C9}
\definecolor{ratinglight}{HTML}{EEF7EE}
\usepackage{graphicx}
\usepackage{booktabs}
\usepackage{array}
\usepackage{amsmath,amssymb}
\usepackage{textcomp}
\usepackage{mathtools}
\usepackage{bm}
\usepackage[italic]{mathastext}
\usepackage{tabularx}
\usepackage{multirow}
\usepackage{makecell}
\usepackage[shortlabels]{enumitem}
\usepackage{microtype}
\usepackage{longtable}
\usepackage{tikz}
\usepackage[numbers,sort&compress]{natbib}
\usepackage{caption}
\usepackage{authblk}

\usepackage[colorlinks=true,linkcolor=M,citecolor=M,urlcolor=M]{hyperref}

\usepackage{titlesec}
\titleformat*{\section}{\large\sffamily\bfseries}
\titleformat*{\subsection}{\normalsize\sffamily\bfseries}
\titleformat*{\subsubsection}{\normalsize\sffamily\bfseries\itshape}
\newcommand{\bmhead}[1]{\paragraph{#1}}

\definecolor{M}{HTML}{1C5985}\definecolor{Z}{HTML}{ECEFF1}\definecolor{muted}{HTML}{9AA0A6}
\newcommand{\chip}[3][2.3em]{\definecolor{CH}{HTML}{#2}\tikz[baseline=-0.6ex]\node[fill=CH,text=white,rounded corners=2pt,inner xsep=3pt,inner ysep=1.4pt,minimum width=#1,font=\sffamily\bfseries\footnotesize]{#3};}
\newcommand{\stripe}[1]{\definecolor{ST}{HTML}{#1}\tikz[baseline=-0.6ex]\fill[ST] (0,-0.17) rectangle (0.06,0.2);}
\newcommand{\rng}[6]{\definecolor{C}{HTML}{#4}\tikz[baseline=-0.55ex]{\path (0,0) -- (#6,0);
  \draw[#5,line width=1.4pt,line cap=round] (#1,0) -- (#2,0);
  \fill[white,draw=#5,line width=0.8pt] (#1,0) circle (0.065);
  \fill[C,draw=white,line width=0.6pt] (#3,0) circle (0.11);}}
\newcommand{\sparknd}[4][2.45]{\definecolor{SC}{HTML}{#4}\tikz[baseline=0.4ex]{\path (0,-0.02) rectangle (#1,0.45);
  \draw[SC,line width=0.8pt,line join=round,line cap=round] #2;}}
\newcommand{\axisw}[1]{\tikz[baseline=-0.55ex]{\path (0,0) -- (#1,0);
  \foreach \v in {1500,1750,2000,2250,2500} \node[font=\sffamily\bfseries\fontsize{6}{6}\selectfont,text=white] at ({(\v-1300)/1350*#1},0) {\v};}}

\definecolor{ZL}{HTML}{F4F6F7}
\newcommand{\hd}[1]{\textcolor{white}{\bfseries #1}}
\newcommand{\sect}[2]{\rowcolor{M!20}\multicolumn{#1}{l}{\bfseries #2}\\}
\newcommand{\tabtop}{\rowcolors{2}{white}{ZL}}
\newcommand{\tabend}{\arrayrulecolor{M}\specialrule{1pt}{0pt}{0pt}}
\newcommand{\totalrule}{\arrayrulecolor{M!40}\specialrule{0.4pt}{1pt}{1pt}}

\begin{document}

\title{\sffamily\bfseries A Modified Glicko Model for Rating Soccer Clubs and National Teams Worldwide}

\author[1]{Andrei Shelopugin\thanks{Corresponding author: \href{mailto:shelopuginandrey@gmail.com}{shelopuginandrey@gmail.com}}}
\author[2]{Alexander Sirotkin\thanks{\href{mailto:avsirotkin@hse.ru}{avsirotkin@hse.ru}}}
\affil[1]{Independent Researcher}
\affil[2]{HSE University}
\date{}

\maketitle

\begin{abstract}
Comparing soccer teams that rarely or never meet is a fundamental challenge in sports analytics: clubs play mostly within their national leagues, and national teams within their confederations. We propose a rating framework that places clubs and national teams worldwide on a common probabilistic scale. Building on a recent extension of Glicko with order effects and strength-dependent draw probabilities, we adapt the model to soccer with goal-scoring adjustments, a decomposition of home advantage, extra time, league-aware rating initialization and rules of promotion and relegation that do not inflate the levels of leagues; national teams start from the levels of their confederations. The models are evaluated on about 1.2 million club matches of 252 leagues from 166 countries (2000--2026) and 12.6 thousand national team matches (1998--2026). On held-out seasons, the modified Glicko model is clearly more accurate than Elo and as accurate as score-matrix models built on the goal expectations of a gradient boosting model, including the Dixon--Coles and Markov chain models, while it is substantially more accurate in international club matches. Averaging the predictions of the rating model and the Markov chain model is more accurate than either model alone. For national teams, the model is significantly more accurate than all baselines, especially in the final tournaments, where teams of different confederations meet. The resulting ratings are interpretable and can be used for player evaluation, transfer analysis and match forecasting. The match data, the ratings and the code are publicly available at \url{https://github.com/andreyshelopugin/SportsRatings}.

\end{abstract}

\section{Introduction}\label{intro}

One of the key challenges for soccer (association football) managers, scouts, and analysts is accurately evaluating player abilities. This can be achieved through video analysis of player performance or by examining performance metrics\cite{bergkamp2022soccer}. In the latter case, it is crucial for analysts to understand the competitive level of different leagues in order to make informed decisions about player recruitment. For example, a player who is the top goal scorer in a lower-tier league may fail to replicate the same performance level after transferring to a more competitive league.

Another important task in soccer analytics is predicting match outcomes~\cite{dixon1997modelling}. To improve the accuracy of match outcome predictions, a ranking model needs to be developed. This model enables the simulation of future match results, helping soccer managers make strategic decisions, particularly regarding player transfers. Additionally, the ranking model can be used to set constraints in draw simulators for final tournaments, ensuring balanced groups and enhancing overall fan engagement.

Some models already calculate ratings of soccer clubs and leagues, such as Club Elo\footnote{\url{http://clubelo.com}, accessed October 30, 2026.} and the UEFA coefficient system, but existing approaches suffer from serious drawbacks. The UEFA rating system\footnote{\url{https://www.uefa.com/nationalassociations/uefarankings/country}, accessed October 30, 2026} is widely regarded in soccer, but it has several weaknesses. Firstly, it does not account for the strength of opponents, treating victories over both top teams and weaker opponents equally. Secondly, it relies heavily on past performances, which can distort a team's current ranking and influence tournament draw outcomes. If a team's performance declines, they may still face stronger opponents, making it difficult to accumulate points. Additionally, the system applies the same weight to matches regardless of the season or year, failing to accurately reflect the current strength of teams.

In addition to these major issues, the UEFA rating system has several minor limitations. For instance, the rating points are not based on a solid mathematical foundation, making the system difficult to interpret. Furthermore, it does not effectively assess the strength of a national league, as it primarily focuses on the performance of top teams, which are typically the ones participating in international competitions. This approach provides little to no information about the performance of weaker teams in the first division or those in lower leagues.

The same problem arises for national teams. The FIFA World Ranking\footnote{\url{https://inside.fifa.com/fifa-rankings/world-ranking/men}, accessed October 30, 2026} has been computed since 2018 with an Elo-type formula~\cite{elo1967proposed}, which accounts for the strength of the opponent. However, it ignores the venue of a match, does not model draws explicitly, and weights matches by fixed importance coefficients of the competitions rather than by coefficients estimated from data. Since most national teams play most of their matches within their own confederation, the levels of the confederations are anchored only by a small number of intercontinental matches, mainly at the World Cup.

This paper aims to address these limitations. We propose a rating system for soccer clubs and national teams based on the Glicko model, with several modifications. The main contributions are as follows:
\begin{itemize}
\item an adaptation of the Glicko model with strength-dependent draws and order effects of Glickman~\cite{glickman2026paired} to soccer: the dependence of the outcome probabilities on the expected number of goals, a decomposition of home advantage into crowd and travel components, and extra time;
\item league-aware rating initialization and rating dynamics at promotion and relegation that do not inflate the levels of leagues, which makes clubs of different leagues and countries comparable;
\item an extension to national teams with confederation starting levels;
\item an evaluation on about 1.2 million matches against the Elo model and against score-matrix models built on the goal expectations of a gradient boosting model: the proposed model is as accurate as the score-matrix models overall and substantially more accurate in international matches, and an ablation attributes this advantage to the league-aware initialization; for national teams, it is significantly more accurate than all baselines, and the levels of the confederations play the same role; together with a comparison with the FIFA ranking;
\item public match data, code and ratings of clubs, leagues and national teams (\url{https://github.com/andreyshelopugin/SportsRatings}), which are planned to be updated regularly.
\end{itemize}

\section{Experimental Setup}

\subsection{Data}

The data were collected from Flashscore\footnote{\url{https://www.flashscore.com}, accessed October 30, 2026} and consist of matches from the first and second divisions of nearly all countries, as well as national cup competitions and international tournaments such as the Champions League and Copa Libertadores. Matches involving teams from lower-ranked leagues (third division and below) were deliberately excluded, including national cup matches featuring third-tier teams. In addition, matches decided by forfeit were removed from the dataset.

The club dataset contains matches played between January 2000 and September 2026 by 8,402 clubs from 252 leagues of 166 countries.

The national team dataset contains matches of 220 national teams played between June 1998 and July 2026: the World Cup and continental final tournaments, their qualifiers starting with the qualifiers for Euro 2000 and the 2002 World Cup, and the Nations Leagues.

The numbers of matches by type of competition are given in Table~\ref{tab:data}.

\begin{table}[!htbp]
\caption{Number of matches by type of competition}
\label{tab:data}
\centering\sffamily\renewcommand{\arraystretch}{1.2}\setlength{\tabcolsep}{8pt}\tabtop
\begin{tabularx}{\textwidth}{X>{\raggedleft\arraybackslash}p{3cm}>{\raggedleft\arraybackslash}p{2.4cm}}
\rowcolor{M}\hd{Competition} & \hd{Matches} & \hd{Share, \%}\\
\sect{3}{Clubs, 2000--2026}
First divisions & 715,021 & 58.9 \\
Second divisions & 408,047 & 33.6 \\
Domestic cups & 55,891 & 4.6 \\
Domestic super cups & 1,530 & 0.1 \\
Continental and intercontinental competitions & 33,606 & 2.8 \\
Continental super cups & 97 & 0.0 \\
\totalrule\bfseries Total & \bfseries 1,214,192 & \bfseries 100.0 \\
\sect{3}{National teams, 1998--2026}
Qualifiers & 9,272 & 73.5 \\
Continental final tournaments & 1,710 & 13.6 \\
World Cup final tournaments & 487 & 3.9 \\
Nations Leagues & 1,149 & 9.1 \\
\totalrule\bfseries Total & \bfseries 12,618 & \bfseries 100.0 \\
\tabend
\end{tabularx}\rowcolors{1}{}{}
\end{table}

The score of a match includes extra time; for 6,491 of the 9,374 matches with extra time, the match pages also give the score after 90 minutes and the goals of extra time separately. Penalty shootouts are not included in the score, and forfeits, walkovers, annulled and abandoned matches are removed. Neutral venues, matches played at the stadium of the visiting team and the hosts of final tournaments are annotated. From the match notes of Flashscore, we also annotate the matches played without spectators (21,549 club matches, mostly during the COVID-19 pandemic), with a limited number of spectators (2,958) and at a different stadium of the home team (2,674); these annotations enter the home advantage (Section~\ref{approach_glicko}).

\subsection{Evaluation Protocol}\label{subsec:protocol}

All models are evaluated on the same held-out seasons. The parameters of all models are estimated on the seasons that started before 2021, and the metrics are computed on the seasons that started in 2021 or later (Table~\ref{tab:split}).

\begin{table}[!htbp]
\caption{Number of matches in the training and test seasons}
\label{tab:split}
\centering\sffamily\renewcommand{\arraystretch}{1.2}\setlength{\tabcolsep}{8pt}\tabtop
\begin{tabularx}{\textwidth}{X>{\raggedleft\arraybackslash}p{2.8cm}>{\raggedleft\arraybackslash}p{2.8cm}>{\raggedleft\arraybackslash}p{2.8cm}}
\rowcolor{M}\hd{Matches} & \hd{Train ($<$ 2021)} & \hd{Test ($\ge$ 2021)} & \hd{Total}\\
Clubs, domestic & 878,816 & 301,756 & 1,180,572 \\
Clubs, international & 24,222 & 9,481 & 33,703 \\
National teams & 8,721 & 3,897 & 12,618 \\
\totalrule\bfseries Total & \bfseries 911,759 & \bfseries 315,134 & \bfseries 1,226,893 \\
\tabend
\end{tabularx}\rowcolors{1}{}{}
\end{table}

 For national teams, the split is by date: the rating models are estimated on the matches played before 2021 and evaluated on the matches since 2021. A split by season would mix the chronology, since a season of national teams spans several years (the qualifiers for the 2022 World Cup started in 2019, and Euro 2020 was played in 2021).
During the test seasons, the parameters are fixed, and the ratings are updated after every match in the order of the dates: the prediction of a match uses the ratings after all earlier matches, as in the use of the model in practice. The predictions are therefore out-of-sample with respect to both the parameters and the results of the matches being predicted. The parameters are not re-estimated during the test seasons, which makes the reported metrics conservative for all models. The national teams are evaluated separately (Section~\ref{subsec:national_results}).

The metrics are computed on the outcome after 90 minutes (win, draw or loss), which is defined for every match. All matches are weighted equally in the test metrics, without the weights and the discounting used in the estimation of the Glicko parameters (Section~\ref{subsec:parameter_estimation}).

The ratings published with this paper are computed with the parameters re-estimated on all seasons, so that the most recent seasons, which matter most for the current strength of the teams, also contribute to the parameters. The two sets of parameters are compared in Section~\ref{subsec:parameters}.

\subsection{Approaches}
There are two effective approaches to predicting soccer match outcomes. The first approach is based on predicting the number of goals each team will score and translating these goal expectations into probabilities of a win, draw, or loss. The second approach relies on rating systems to quantify team strength, in which the probabilistic nature is embedded in the model formulation. Consequently, probabilities of a win, draw, or loss arise directly from the rating dynamics. We discuss both approaches in sections \ref{approach_poisson} and \ref{approach_glicko}, respectively.

\section{Approach 1: Match Outcome Prediction via Goal Expectations}\label{approach_poisson}
\subsection{Idea}\label{subsec21}

This approach leverages a widely recognized concept for forecasting soccer match results, focusing on predicting the number of goals each team will score against its respective opponent. We assume that the number of goals in soccer \textit{approximately} follows a Poisson distribution~\cite{dixon1997modelling}. We then translate these goal expectations into outcome probabilities.

\subsection{Implementation}
Gradient boosting models have demonstrated impressive performance when applied to tabular data \cite{grinsztajn2022tree}. Consequently, we constructed a machine learning model utilizing the LightGBM \cite{ke2017lightgbm} framework to predict the goal count for both competing teams. We also experimented with the CatBoost \cite{prokhorenkova2018catboost} model; however, it yielded inferior results. Since the goals in soccer approximately follow a Poisson distribution, we set the LightGBM parameter \texttt{objective} to \texttt{poisson}. We note, however, that empirical goal distributions deviate from the strict Poisson assumption. This limitation is addressed later through model refinements.

The goal models (LightGBM and CatBoost, with the same features) predict the expected number of goals of each team in a match. Each row of the training set corresponds to one participant of a match, so every match gives two rows. The model uses 52 features listed in Table~\ref{tab:features}. The rolling statistics are computed over the last 5, 10, 20 and 30 matches of the team (one feature per window) with at least 3, 5, 10 and 20 matches, respectively; otherwise the feature is missing.

Matches with extra time last 120 minutes instead of 90. Where the score after 90 minutes is known, the goal model is trained on it. For the other matches with extra time, the final score is used with an offset: the model expects $1.38$ times more goals than in 90 minutes. The ratio $0.38$ of the goals of extra time to the goals of a regular match is estimated from the data: 1.02 goals are scored in extra time against 2.69 in an average match (12,898 matches with extra time and the scores by period, including the lower divisions and cups that are not part of the dataset). Per minute, more goals are scored in extra time than in an average match (3.05 against 2.69 per 90 minutes). The prediction for every match is the expected number of goals in 90 minutes, and the expected number of goals in extra time is $0.38$ of it.

\begin{table}[!htbp]
\caption{Features of the gradient boosting models (LightGBM and CatBoost) that predict the expected number of goals of a team in a match. Every row of the training set is a pair (match, team), so every match gives two rows. Venue: in the last home matches for a home match and in the last away matches for an away match}
\label{tab:features}
\centering\sffamily\renewcommand{\arraystretch}{1.2}\tabtop
\begin{tabularx}{\textwidth}{p{2.6cm}X}
\rowcolor{M}\hd{Group} & \hd{Features}\\
\bfseries Venue & venue (+1 home, 0 neutral, \textminus 1 away); no spectators; limited number of spectators; home match at a different stadium; match during the COVID-19 pandemic\\
\bfseries Stage & extra time possible\\
\bfseries Schedule & days since the previous match; matches played in the season; both for the team and for the opponent\\
\bfseries Competition & league of the team and of the opponent; tournament; tournament type (first league, second league, cup, super cup, international cup, international super cup, national team qualification, national team final tournament); league level difference (+1, 0, \textminus 1)\\
\end{tabularx}\rowcolors{1}{}{}

\vspace{8pt}
\renewcommand{\arraystretch}{1.2}\tabtop
\begin{tabularx}{\textwidth}{p{2.6cm}X>{\centering\arraybackslash}p{1.15cm}>{\centering\arraybackslash}p{1.15cm}>{\centering\arraybackslash}p{1.15cm}>{\raggedleft\arraybackslash}p{2.4cm}}
\rowcolor{M}\hd{Group} & \hd{Rolling statistic} & \hd{Mean} & \hd{Median} & \hd{Max} & \hd{Windows}\\
\bfseries Form & goals scored & $\bullet$ & & & 5, 10, 20, 30\\
 & goals conceded & $\bullet$ & & & 5, 10, 20, 30\\
 & goals scored by the opponent & $\bullet$ & & & 5, 10, 20, 30\\
 & goals conceded by the opponent & $\bullet$ & & & 5, 10, 20, 30\\
 & goals scored, same tournament type & $\bullet$ & & & 5, 10, 20, 30\\
\bfseries Venue form & goals scored at the venue & $\bullet$ & $\bullet$ & $\bullet$ & 5, 10, 20, 30; max: 10\\
 & goals conceded at the venue & $\bullet$ & $\bullet$ & & 5, 10, 20, 30\\
\tabend
\end{tabularx}\rowcolors{1}{}{}
\end{table}

\subsection{Outcome Probabilities via the Skellam Distribution}

The Poisson assumption, together with the assumption of independence between the goal-scoring processes of the two teams, allows us to use the probability mass function associated with the Skellam distribution:

\begin{equation}
    p(k, \mu_1, \mu_2) = exp(-(\mu_1 + \mu_2)) (\frac{\mu_1}{\mu_2})^{\frac{k}{2}}I_k(2\sqrt{\mu_1 \mu_2})
\label{skellam}
\end{equation}

In equation \eqref{skellam}, $I_k$ is the modified Bessel function, and $k$ represents the score difference in a soccer match. If $k > 0$, the value of the probability mass function corresponds to the probability of the first team winning; for $k = 0$, it gives the draw probability; and if $k < 0$, it provides the probability of the second team winning. The parameters $\mu_1$ and $\mu_2$ are the means of two Poisson-distributed random variables. In our case, $\mu_1$ and $\mu_2$ are the expected goals of the two teams predicted by the gradient boosting model.

\subsection{Score Matrix Models}

While the Skellam distribution provides a direct way to compute win, draw, and loss probabilities from goal expectations, it does not explicitly represent the full joint distribution of possible scorelines. To obtain a more detailed description of match outcomes, we construct a score probability matrix.

Let $p(i,j)$ denote the probability that the home team scores $i$ goals and the away team scores $j$ goals. The collection of probabilities $\{p(i,j)\}$ forms a two-dimensional matrix representing the joint distribution of match scores. In practice, the infinite support of goal counts must be truncated for computational purposes. We approximate the joint distribution by constructing a $15 \times 15$ score matrix corresponding to goal counts from $0$ to $14$ for each team. For typical goal expectations observed in professional soccer, the probability of exceeding this range is negligible; the matrix is subsequently renormalized so that the total probability mass equals one.

Outcome probabilities are then computed by summing over appropriate regions of this matrix:

\begin{equation}
\begin{aligned}
p_H &= \sum_{i>j} p(i,j), \\
p_D &= \sum_{i=j} p(i,j), \\
p_A &= \sum_{i<j} p(i,j).
\end{aligned}
\label{eq:matrix_outcomes}
\end{equation}

This matrix-based formulation offers several advantages. First, it provides the complete distribution over scorelines rather than only the score difference. Second, it allows flexible modeling of dependence structures between the two teams' goal processes.
Third, the outcome probabilities of all score-matrix models below are computed in the same way, by Eq.~\eqref{eq:matrix_outcomes}, so the models differ only in how the matrix is constructed.

In the following subsections, we describe three alternative constructions of the score matrix.

\subsection{Score Matrix under Independent Poisson Assumption}

The simplest approach assumes that the number of goals scored by the home and away teams follows independent Poisson distributions with means $\mu_1$ and $\mu_2$, respectively.

Let $x_i = p(G_{\text{home}}=i)$ and
$y_j = p(G_{\text{away}}=j)$ denote the corresponding Poisson probabilities. Under the independence assumption, the joint score probability is given by

\begin{equation}
p(i,j) = x_i y_j.
\label{eq:independent_poisson_matrix}
\end{equation}

\subsection{Dixon--Coles Score Matrix}

Although the independent Poisson model provides a convenient baseline, empirical soccer data reveal dependence between the goal-scoring processes of the two teams, particularly in low-scoring matches. The independent model tends to underestimate the probabilities of low-scoring draws such as $(0,0)$ and $(1,1)$. To account for this, we adopt the Dixon--Coles adjustment~\cite{dixon1997modelling}. Starting from the independent Poisson construction given in Equation~\eqref{eq:independent_poisson_matrix}, the probabilities of specific low-score outcomes are modified.

The Dixon--Coles model introduces a dependence parameter $\rho$ that adjusts the probabilities as follows:

\begin{equation}
\begin{aligned}
p(0,0) &\leftarrow p(0,0)\,(1 - \mu_1 \mu_2 \rho), \\
p(0,1) &\leftarrow p(0,1)\,(1 + \mu_1 \rho), \\
p(1,0) &\leftarrow p(1,0)\,(1 + \mu_2 \rho), \\
p(1,1) &\leftarrow p(1,1)\,(1 - \rho).
\end{aligned}
\label{eq:dixon_coles_adjustment}
\end{equation}

All other scoreline probabilities remain unchanged. After applying the correction factors, the score matrix is renormalized to ensure that the total probability sums to one.

When $\rho = 0$, the model reduces to the independent Poisson formulation. The parameter is estimated on the validation set of the goal model by a grid search with the step 0.001 (Section~\ref{subsec:score_models}): $\rho = -0.060$.

\subsection{Markov Score Matrix}

The Poisson and Dixon--Coles matrices describe only the final score. In a real match, however, the scoring rate of a team depends on the current score: a leading team tends to defend, and a trailing team takes more risk. Following the birth process model of Dixon and Robinson~\cite{dixon1998birth}, we describe a match as a sequence of goals: the score starts at 0--0, and from every score each team scores at its own rate, which depends on the goal difference. Unlike the original model, which estimates the strength of the teams itself and is fitted on the times of the goals, we use it as a layer on top of the goal model: the base rates of the two teams are the expected goals of LightGBM, and only the dependence on the score is estimated, from the final outcomes.

From a score $(i, j)$ the home team scores with the rate $\lambda_{\text{home}}(i, j)$ and the away team with the rate $\lambda_{\text{away}}(i, j)$:
\begin{equation}
\begin{aligned}
\lambda_{\text{home}}(i, j) &= \mu_{\text{home}} \, m_{\text{score}}(d) \, m_{\text{low}}(i, j), \\
\lambda_{\text{away}}(i, j) &= \mu_{\text{away}} \, m_{\text{score}}(-d) \, m_{\text{low}}(i, j),
\end{aligned}
\label{eq:state_dependent_rates}
\end{equation}
where $d = i - j$ is the goal difference, capped at $\pm 3$, $m_{\text{score}}(d)$ is the multiplier of a team leading by $d$ goals ($d < 0$: trailing by $|d|$ goals), $m_{\text{score}}(0) = 1$, and $m_{\text{low}}(i, j)$ adjusts the low scores in the spirit of Dixon--Coles, scaled by $\mu_{\text{home}} \mu_{\text{away}}$:
\begin{equation}
\begin{aligned}
m_{\text{low}}(0, 0) &= 1 + a_{00} \mu_{\text{home}} \mu_{\text{away}}, \\
m_{\text{low}}(1, 0) = m_{\text{low}}(0, 1) &= 1 - a_{10} \mu_{\text{home}} \mu_{\text{away}}, \\
m_{\text{low}}(1, 1) &= 1 + a_{11} \mu_{\text{home}} \mu_{\text{away}},
\end{aligned}
\end{equation}
$m_{\text{low}}(i, j) = 1$ for the other scores. The score matrix $p(i, j)$ is the distribution of the state of the chain at the end of the match, the matrix exponential of the generator (transition rate) matrix built from these rates. With $m_{\text{score}} \equiv 1$ and $m_{\text{low}} \equiv 1$ the chain reduces to the independent Poisson matrix.

The multipliers change not only the shape of the score distribution but also the total number of goals. Since the expected goals are taken from LightGBM, we rescale the base rates so that the chain produces the same expected goals as LightGBM, and the Markov layer changes only the shape of the distribution.

The nine parameters are estimated on the validation set of the goal model (Section~\ref{subsec:score_models}) by minimizing the outcome log loss; Table~\ref{tab:markov} shows the estimates.
A one-goal lead makes the leading team score less and the trailing team score more, a strong comeback effect, in agreement with Dixon and Robinson~\cite{dixon1998birth}; with a three-goal lead the effect reverses (Table~\ref{tab:markov}). Part of the reason is that the match is effectively decided, but a three-goal lead is also a selected state: it often arises when something went wrong for the trailing team, such as a red card or a collapse of the team, which is not in the expected goals before the match and keeps acting until the end of the match.

The estimates should be read as the direction and the approximate size of the effects rather than as precise values. The loss changes very little with the parameters, since different combinations of them give almost the same probabilities of a win, a draw and a loss, and large goal differences are rare. With the minutes of the goals, as used by Dixon and Robinson~\cite{dixon1998birth}, the time spent at every score and the goals scored from it would be observed directly, and the multipliers would be identified much better.

\begin{table}[!htbp]
\caption{Fitted parameters of the Markov score matrix, Eq.~\eqref{eq:state_dependent_rates}: the scoring rates of the leading and the trailing team relative to the base rates of the goal model, and the low-score adjustments}
\label{tab:markov}
\centering\sffamily\renewcommand{\arraystretch}{1.2}\tabtop
\begin{tabularx}{\textwidth}{>{\centering\arraybackslash}p{1.6cm}>{\raggedleft\arraybackslash}p{1.7cm}>{\raggedleft\arraybackslash}p{1.5cm}>{\raggedleft\arraybackslash}p{1.7cm}>{\raggedleft\arraybackslash}p{1.5cm}X}
\rowcolor{M}\hd{Goal} & \multicolumn{2}{c}{\hd{Leading team}} & \multicolumn{2}{c}{\hd{Trailing team}} & \\
\rowcolor{M}\hd{difference} & \hd{Multiplier} & \hd{Change} & \hd{Multiplier} & \hd{Change} & \hd{Interpretation} \\
1 & 0.800 & \textminus 20.0\% & 1.205 & +20.5\% & leader defends, trailer attacks \\
2 & 0.960 & \textminus 4.0\% & 0.930 & \textminus 7.0\% & both teams score less \\
3 & 1.230 & +23.0\% & 0.855 & \textminus 14.5\% & match decided; red cards \\
\sect{6}{\mdseries Low-score adjustments: $a_{00} = 0.025$, $a_{10} = 0.085$, $a_{11} = 0.040$}
\tabend
\end{tabularx}\rowcolors{1}{}{}
\end{table}

\section{Approach 2: Match Outcome Prediction Using Modified Glicko}\label{approach_glicko}

\subsection{Rating Systems}

Rating systems such as Elo \cite{elo1967proposed}, Glicko \cite{glickman2012example}, and TrueSkill \cite{herbrich2006trueskill} have been widely applied to measure competitor strength in head-to-head games and sports. These systems were originally designed for competitions with binary outcomes (win or loss). Most implementations treat ties as equivalent to half a win and half a loss, ignoring the fact that draw probabilities may depend not only on the difference in competitor strengths. However, their direct application to team sports such as soccer presents significant challenges, as standard rating models do not account for crucial sport-specific factors, including home-field advantage, roster changes due to player transfers, and strength-dependent draw probabilities.

In soccer, the occurrence of draws introduces additional complexity that classical rating systems cannot naturally accommodate. Empirical evidence suggests that draw probabilities depend not only on the relative strength difference between teams, but also on their absolute strength levels and goal-scoring tendencies. Recently, Glickman~\cite{glickman2026paired} extended paired comparison models to incorporate strength-dependent tie probabilities in chess, recognizing that draws become more likely when competitors are closely matched in skill, and that stronger players are more likely to draw than weaker players. This development provides a principled foundation for adapting rating systems to soccer, where draws are similarly common and influenced by both team strengths.

\begin{figure}[!h]
\centering
\includegraphics[width=\textwidth]{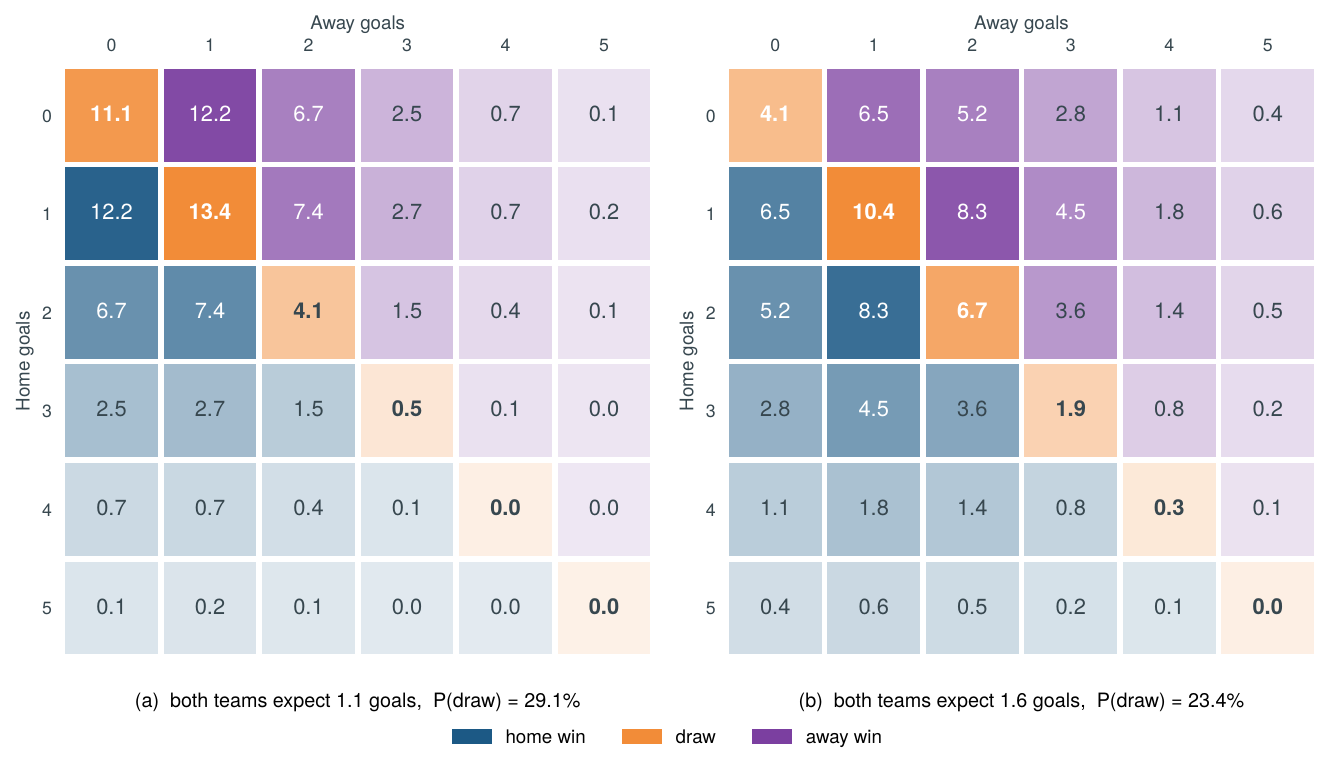}
\caption{Score probabilities (\%) for two equally strong teams under independent Poisson goals: (a) both teams expect 1.1 goals, (b) both teams expect 1.6 goals. Home wins are shown in blue, draws (the diagonal) in orange and away wins in purple, darker for more probable scores. The teams are equal in both cases, but the draw probability falls from 29.1\% to 23.4\% as the expected number of goals grows.}\label{fig:draws}
\end{figure}

In our previous work~\cite{shelopugin2023ratings}, we adapted the Glicko-2 model to soccer by incorporating draw probabilities, home-field advantage, and rating adjustments for league transitions. Building on these developments and Glickman's recent advances, we now adopt the strength-dependent draw framework from~\cite{glickman2026paired}, extending it to account for goal-scoring tendencies—a soccer-specific factor where high-scoring teams produce fewer draws than low-scoring teams (Fig.~\ref{fig:draws}). This relationship is consistent with the nature of goal-scoring in soccer: teams with lower expected goal rates are more likely to remain at 0-0, thereby increasing the overall probability of a draw, while high-scoring teams generate more total goals, reducing the likelihood that both teams finish with identical scores.

Furthermore, we incorporate league-specific characteristics and dynamic rating adjustments following promotion or relegation. Our approach enables cross-league comparisons and provides interpretable ratings that can be applied to player evaluation, transfer analysis, and match outcome prediction across diverse competitive environments.

\subsection{Glicko with Strength-Dependent Draws}

The Glicko model represents competitor strength as a probability distribution. Each team is initially associated with a normal prior distribution characterized by a mean rating $\mu$ and uncertainty (rating deviation) $\phi$. As match outcomes accumulate, ratings are updated via Bayesian inference toward a posterior distribution. When a team wins, its rating increases; the magnitude of this increase is primarily determined by two factors: the difference in ratings (larger upsets produce larger adjustments) and rating uncertainty (teams with limited match history experience larger changes).

Recent work by Glickman~\cite{glickman2026paired} extends paired comparison models to address two critical limitations of classical rating systems. First, order effects such as home-field advantage (analogous to the white piece advantage in chess) systematically favor one competitor. Second, the probability of a draw depends not only on the strength difference between competitors, but also on their absolute strength levels—stronger pairs are more likely to draw.

The original formulas are given below:

\begin{align}
P(Y_{ij} = 1) &= \frac{\exp\left( \theta_i + \frac{1}{4} x_{ij} \left( \alpha_0 + \alpha_1 \cdot \frac{\theta_i + \theta_j}{2} \right) \right)}{D_{ij}} \label{eq:win_prob}\\
P(Y_{ij} = 0) &= \frac{\exp\left( \theta_j - \frac{1}{4} x_{ij} \left(  \alpha_0 + \alpha_1 \cdot \frac{\theta_i + \theta_j}{2} \right) \right)}{D_{ij}} \label{eq:loss_prob}\\
P(Y_{ij} = \tfrac{1}{2}) &= \frac{\exp\left( \beta_0 + (1 + \beta_1) \cdot \frac{\theta_i + \theta_j}{2} \right)}{D_{ij}} \label{eq:draw_prob}
\end{align}

where
\begin{itemize}[label={--}, nosep, leftmargin=*]
\item $\theta_i$, $\theta_j$ are the ratings of team $i$ (home) and team $j$ (away);
\item $x_{ij}$ is the venue: $1$ home, $0$ neutral, $-1$ away for team $i$;
\item $\alpha_0$, $\alpha_1$ are the baseline and the strength-dependent home advantage;
\item $\beta_0$, $\beta_1$ are the baseline and the strength-dependent draw probability;
\item $D_{ij}$ is the sum of the three numerators.
\end{itemize}

The term $\frac{\theta_i + \theta_j}{2}$ represents the average strength of both teams. In Equations~\eqref{eq:win_prob}--\eqref{eq:loss_prob}, this term modulates the home-field advantage: the parameter $\alpha_1$ allows the advantage to vary with team quality. In Equation~\eqref{eq:draw_prob}, the coefficient $(1 + \beta_1)$ determines how draw probability changes with absolute team strength.

\subsection{Custom Extensions of Glicko-2}
\subsubsection{Modified Outcome Probability Function}

We modify the original probability specification by (i) incorporating expected goals into both the order effect and the draw component, and (ii) decomposing the baseline order effect into a crowd-related term and a logistic (travel) term.
Let $g_{ij}$ denote the scoring intensity of the match: the sum of the expected goals of the two teams from the goal model (Section~\ref{approach_poisson}), clipped at its 99.9\% quantile for numerical stability and centered at its mean over the club matches.

The match outcome probabilities are defined as:

\begin{align}
P(Y_{ij} = 1) &=
\frac{
\exp\left(
\theta_i
+
x_{ij}
\left(
\alpha_{\mathrm{crowd}}
+
\alpha_{\mathrm{logistic}}
+
\alpha_1 \frac{\theta_i + \theta_j}{2}
+
\alpha_2 g_{ij}
\right)
\right)
}{D_{ij}},
\\[8pt]
P(Y_{ij} = 0) &=
\frac{
\exp\left(
\theta_j
-
x_{ij}
\left(
\alpha_{\mathrm{crowd}}
+
\alpha_{\mathrm{logistic}}
+
\alpha_1 \frac{\theta_i + \theta_j}{2}
+
\alpha_2 g_{ij}
\right)
\right)
}{D_{ij}},
\\[8pt]
P(Y_{ij} = \tfrac{1}{2}) &=
\frac{
\exp\left(
\beta_0
+
\beta_1 \frac{\theta_i + \theta_j}{2}
+
\beta_2 g_{ij}
\right)
}{D_{ij}}.
\end{align}

Here, $x_{ij}$ indicates venue, $\alpha_{\mathrm{crowd}}$ and $\alpha_{\mathrm{logistic}}$ jointly determine the baseline home advantage, $\alpha_1$ captures strength-dependent order effects, and $\alpha_2$ introduces a scoring-intensity adjustment. The draw component is governed by $\beta_0$ (baseline draw frequency), $\beta_1$ (strength dependence), and $\beta_2$ (scoring dependence).

The separation between $\alpha_{\mathrm{crowd}}$ and $\alpha_{\mathrm{logistic}}$ is identifiable due to the COVID-19 period, during which matches were played without spectators. While the crowd-related component was effectively removed, travel-related disadvantages remained, providing natural variation that allows the two effects to be estimated separately. Beyond the pandemic period, match-level annotations from Flashscore—such as ``Limited number of spectators'' or ``Playing home matches at a different stadium''—enable further fine-tuning of these parameters.
Specifically, the crowd component is set to zero for matches without spectators and halved for matches with a limited number of spectators; for a home match at a different stadium, the crowd component is halved and the travel component is set to zero, since in this case both teams usually travel to the venue (a simplification: the alternative stadium is sometimes close to the home city). At a neutral venue, the home advantage is zero, and the draw component uses the average of the parameters of the leagues of the two teams.

\subsubsection{Rating Initialization}

The Glicko model depends only on rating differences; adding a constant to all ratings does not affect predicted match outcomes. To enable cross-league comparability, we introduce an additional parameter referred to as the \emph{initial league rating}, which represents the average strength level of a league. While this parameter does not affect predictions within a single national championship, it plays a crucial role when comparing teams across different leagues, particularly in international competitions.

Every club starts with the initial rating of its league, $\mu = \mu_{\text{init}}$, and the initial rating deviation $\phi_0 = 250$. The initial league rating is used only when the league itself appears in the dataset (or the club appears in its first season in the dataset). Later, the initial rating of a club that enters a league is determined by the clubs it replaces, which keeps the level of the league stable (Section~\ref{subsec:invariant}):
\begin{equation}
\mu =
 \begin{cases}
   \mu + \mu_{\text{lift}} \ \text{or} \ \mu - \mu_{\text{lift}} & \text{if the club was promoted or relegated (Eq.~\eqref{eq:tied_lift})}, \\
   \overline{\mu}_{\text{leavers}} + \mu_{\text{new}} & \text{if the club has no matches in the previous season}, \\
   \mu_{\text{init}} & \text{in the first season of the league}.
 \end{cases}
\label{rating_init_formula}
\end{equation}
A club without matches in the previous season is reinitialized even if earlier history exists: for example, a club that played in the second division in 2013/2014, was relegated to the third division (not in the dataset) and returned in 2015/2016 starts the 2015/2016 season from the anchor of Eq.~\eqref{eq:newcomer_anchor}, and its rating deviation is reset to $\phi_0$.

For leagues that appear only after the training seasons, the parameters are the medians over the fitted first or second divisions; for leagues of countries with fewer than 30 international matches, the initial rating is additionally capped at 1400, below the mean rating of 1500: their level cannot be identified from the data, and leagues with little international history are, as a rule, weaker than average.

\subsubsection{Mid-Season}

As in Glicko, the matches are grouped into rating periods: the ratings and rating deviations are updated once per period, using all matches of the period together. The rating periods are the calendar quarters, which gives 107 periods between 2000 and 2026 and about ten matches of a team per period in a typical league. The length of the period is a hyperparameter: shorter periods update the ratings more often but with less information per update, longer periods react more slowly to changes of form. For the predictions, the ratings within a period are updated after every match in the order of the dates (Section~\ref{subsec:protocol}), so the length of the period affects the official update of the ratings but not the timeliness of the predictions.

Team strength changes over time, gradually during a season and abruptly at the season transition, when the rosters change in the transfer window. Both are modeled by the growth of the rating deviation. Before every rating period (a quarter of a year), the rating deviation of a team grows as in Glicko,
\begin{equation}
\phi \leftarrow \sqrt{\phi^2 + \phi_{\text{period}}^2},
\label{eq:rd_period}
\end{equation}
as long as $\phi$ is below the threshold of 120. At the season transition, it additionally grows by $\phi_{\text{season}}$, and a team that changes its division also changes its rating by $\mu_{\text{lift}}$:
\begin{equation}
r(\mu,\phi) \leftarrow
 \begin{cases}
   r(\mu \pm \mu_{\text{lift}},\ \phi + \phi_{\text{season}}) &\text{if the team was promoted ($+$) or relegated ($-$)},\\
   r(\mu,\ \phi + \phi_{\text{season}}) &\text{otherwise.}
 \end{cases}
 \label{mid_season_formula}
\end{equation}
The parameter $\mu_{\text{lift}} \ge 0$ is estimated for every country and captures the tendency of promoted clubs to strengthen (the budget typically increases after promotion) and of relegated clubs to weaken. The growth $\phi_{\text{season}}$ has two global values, one for the first and one for the second divisions, since the rosters of second-division clubs change more between seasons (per-league values did not improve the model, Section~\ref{subsec:negative}); the estimates are given in Section~\ref{subsec:parameters}.

Another source of inflation is the change in the number of teams of a league, for example after a reform of its format. It is handled by the same mechanism: the newcomers start from the ratings of the clubs they replace, or, in an expansion of the league, from a low quantile of the ratings of the continuing clubs (Section~\ref{subsec:invariant}).

\subsubsection{Preventing Rating Inflation at League Transitions}\label{subsec:invariant}

Matches within a league are almost zero-sum: the rating gained by one team is lost by its opponent. The levels of the leagues are therefore determined by two sources: international matches and the rating that enters or leaves a league at the season transitions. If the transitions systematically add rating to a league, the league is inflated, and since most leagues play few international matches, the inflation is not corrected by the data.

In the earlier version of the model~\cite{shelopugin2026constructing}, promotion and relegation had separate parameters, estimated independently for every league. Such parameters are poorly identified within a single league, and the estimates of some leagues added more rating to promoted teams than was taken from relegated teams. As a result, the total rating of these leagues grew from season to season, and their clubs were overrated in international competitions. The imbalance of a single transition is small, but it is repeated at every season transition, and over the 25 seasons of the dataset it accumulates into a noticeable drift of the level of the league. The effect resembles exploding and vanishing gradients in deep and recurrent neural networks, where a small deviation at every step is compounded over a long sequence of steps. A similar imbalance exists in the FIFA ranking of national teams: matches decided by a penalty shoot-out are scored as 0.75 for the winner and 0.5 for the loser, so such a match adds 1.25 instead of 1 to the total score of the two teams, and the teams that often reach the knockout stages accumulate points in excess of their results.

In the proposed model, the transition between the first and the second division of a country is governed by a single parameter $\mu_{\text{lift}} \ge 0$, estimated for the first division: a promoted team gains $\mu_{\text{lift}}$, and a relegated team loses the same amount,
\begin{equation}
\mu \leftarrow \mu + \mu_{\text{lift}} \ \text{(promotion)}, \qquad \mu \leftarrow \mu - \mu_{\text{lift}} \ \text{(relegation)}.
\label{eq:tied_lift}
\end{equation}
When the numbers of promoted and relegated teams are equal, the transitions do not change the total rating of the two divisions of the country.

A team that enters a league without a match history in the previous season (for example, after a few seasons in the third division, which is not in the dataset) starts from the mean current rating of the teams that left the league at the same transition, plus a global offset $\mu_{\text{new}}$:
\begin{equation}
\mu = \overline{\mu}_{\text{leavers}} + \mu_{\text{new}}.
\label{eq:newcomer_anchor}
\end{equation}
If the league has no leavers (e.g. an expansion of the league), the starting rating is the 10\% quantile of the ratings of the continuing teams, and in the first season of a league, the initial league rating $\mu_{\text{init}}$. The offset is positive ($\mu_{\text{new}} = 90$): at the moment of leaving, the ratings of the leavers are too low because of regression to the mean and the collapse of some clubs (financial problems, the departure of key players, administrative relegation), so the newcomers start above them. Since a newcomer replaces a leaver, the total rating of the league changes only by the offset. One value is used for all leagues (Section~\ref{subsec:negative}).

The offset slightly increases the total rating of a league at every transition. We also tested a stricter version, in which the continuing teams of the league give up the same total amount, so the total rating of the league is preserved exactly. This version made the predictions worse for the newcomers, the relegated teams and the international matches (Section~\ref{subsec:calibration}), so it is not used. The reason is that exact conservation is only needed where nothing else anchors the level of a league. In leagues connected to other leagues by international matches, the ratings of the continuing teams are already corrected by these matches, and an additional downward shift only distorts them. The small inflation from the newcomer offset matters only for isolated leagues with few international matches, whose level is in any case less reliable (Section~\ref{subsec:league_ratings}). The goal of the model is therefore not an exactly constant total rating, but the absence of a systematic drift that the data cannot correct: the large imbalance of separate promotion and relegation parameters is removed, while the small, justified offset of the newcomers is kept.

\subsubsection{Extra Time}\label{subsec:extra_time}

In single-leg knockout matches and in the second legs of two-legged ties with a level aggregate score, a draw after 90 minutes is often followed by extra time. Such matches are rare (9,122 club matches, 0.75\% of the dataset, and 252 national team matches, 2.0\%), but they are the playoffs of the most important tournaments, where an accurate assessment of the strongest teams matters most. We treat extra time as a second stage of the match with its own outcome (win, draw or loss in extra time), which enters both the rating update and the loss function in the same way as the outcome of the regular time. Penalty shootouts directly after 90 minutes, in competitions without extra time (2,142 matches, identified by the scores by period), are draws of the regular time without the extra time stage.

Extra time is described by the same outcome model as the regular time, applied to a shorter game, rather than by a separate model. The only question is how the parameters of the outcome model change when the game lasts 30 minutes instead of 90. A shorter game contains fewer goals, so the outcome depends less on the difference in strength and more on chance. In the model, this corresponds to a contraction of the rating scale by a factor $\kappa$: the ratings enter the outcome probabilities as $\kappa\theta$, and $\kappa = 1$ recovers the regular time. Similarly, the home advantage of the league is scaled by a factor $s_{\mathrm{ET}}$, and the draw component, which describes the probability that 30 minutes end without a decisive goal, has its own parameters $\beta_0^{\mathrm{ET}}, \beta_1^{\mathrm{ET}}, \beta_2^{\mathrm{ET}}$. All five parameters are common for all leagues, since extra time is played in only a few matches per league, and they are estimated from the data together with the other parameters of the model:
\begin{align}
P(Y^{\mathrm{ET}}_{ij} = 1) &\propto \exp\left( \kappa\theta_i + x_{ij}\, s_{\mathrm{ET}} \left( \alpha_{\mathrm{crowd}} + \alpha_{\mathrm{logistic}} + \alpha_1 \tfrac{\theta_i + \theta_j}{2} + \alpha_2 g^{\mathrm{ET}}_{ij} \right) \right), \\
P(Y^{\mathrm{ET}}_{ij} = 0) &\propto \exp\left( \kappa\theta_j - x_{ij}\, s_{\mathrm{ET}} \left( \alpha_{\mathrm{crowd}} + \alpha_{\mathrm{logistic}} + \alpha_1 \tfrac{\theta_i + \theta_j}{2} + \alpha_2 g^{\mathrm{ET}}_{ij} \right) \right), \\
P(Y^{\mathrm{ET}}_{ij} = \tfrac{1}{2}) &\propto \exp\left( \beta_0^{\mathrm{ET}} + \beta_1^{\mathrm{ET}} \kappa \tfrac{\theta_i + \theta_j}{2} + \beta_2^{\mathrm{ET}} g^{\mathrm{ET}}_{ij} \right),
\end{align}
where $g^{\mathrm{ET}}_{ij}$ is the expected number of goals in extra time: the expected number of goals of 90 minutes multiplied by $0.38$ (Section~\ref{approach_poisson}). A match with extra time thus acts as two games: a draw of the regular time, which updates the ratings as any other draw, and a short second game, whose result moves the ratings $\kappa$ times less than an equally surprising result of a regular match.
The parameters of extra time are estimated after the parameters of the leagues; the estimates are given in Table~\ref{tab:parameters}.

With the estimated parameters ($\kappa = 0.76$, $s_{\mathrm{ET}} = 0.58$), a draw is by far the most likely outcome of extra time, and the difference in strength and the home advantage matter much less than in regular time (Table~\ref{tab:interpretation}).

\subsection{Parameter Estimation}\label{subsec:parameter_estimation}

The parameters of the model are listed in Table~\ref{tab:parameters}. They are of two kinds: the parameters of a league are estimated separately for every first and second division, and the global parameters are common for all leagues. The transition parameter $\mu_{\text{lift}}$ is estimated for the first division of every country and applies to both of its divisions. The offset of the newcomers is not estimated but set to $\mu_{\text{new}} = 90$: newcomers play a small share of the matches, so the loss hardly changes with this value. Instead, the value is checked by the calibration: in their first season in the league, the newcomers score as expected by the model ($+0.006$ per match, Table~\ref{tab:calibration}). The table gives the estimates on all seasons, which are used for the published ratings; the estimates on the training seasons, used for the evaluation, are close to them (Section~\ref{subsec:parameters}).

\begin{table}[!htbp]
\caption{Parameters of the model and their estimates on all seasons. League: estimated for every league (253 leagues: 165 first and 88 second divisions); country: for every country (165 countries); global: once for all leagues; fixed: set without estimation by the model. For the league and country parameters, the mean over the leagues and, in brackets, the standard deviation}
\label{tab:parameters}
\centering\sffamily\renewcommand{\arraystretch}{1.2}\tabtop
\begin{tabularx}{\textwidth}{p{2.4cm}>{\raggedleft\arraybackslash}p{2.8cm}@{\hspace{2em}}X}
\rowcolor{M}\hd{Parameter} & \hd{Estimate} & \hd{Description}\\
\sect{3}{League}
$\mu_{\text{init}}$ & 1330 (253) & initial rating of the league \\
$\alpha_{\mathrm{crowd}}$ & 0.101 (0.068) & home advantage: crowd \\
$\alpha_{\mathrm{logistic}}$ & 0.119 (0.067) & home advantage: travel \\
$\alpha_1$ & 0.004 (0.009) & home advantage: strength of the teams \\
$\alpha_2$ & \textminus 0.022 (0.019) & home advantage: expected goals \\
$\beta_0$ & \textminus 0.245 (0.069) & draws: baseline \\
$\beta_1$ & 1.009 (0.012) & draws: strength of the teams \\
$\beta_2$ & \textminus 0.495 (0.023) & draws: expected goals \\
\sect{3}{Country}
$\mu_{\text{lift}}$ & 37.6 (12.9) & rating change at promotion and relegation \\
\sect{3}{Global}
$\phi_{\text{period}}$ & 14 & growth of $\phi$ per rating period \\
$\phi_{\text{season}}$ & 16 & growth of $\phi$ per season, first divisions \\
$\phi_{\text{season}}$ & 34 & growth of $\phi$ per season, second divisions \\
$\kappa$ & 0.76 & extra time: scale of the ratings \\
$s_{\mathrm{ET}}$ & 0.58 & extra time: scale of the home advantage \\
$\beta^{\mathrm{ET}}_0$ & 1.01 & extra time: draws, baseline \\
$\beta^{\mathrm{ET}}_1$ & 1.00 & extra time: draws, strength of the teams \\
$\beta^{\mathrm{ET}}_2$ & \textminus 1.63 & extra time: draws, expected goals \\
\sect{3}{Fixed}
$\mu_{\text{new}}$ & 90 & rating offset of the newcomers \\
 & 4 & rating periods per year (calendar quarters) \\
$\gamma$ & 0.95 & discount of the loss per rating period \\
$w$ & 4 & loss weight: continental club competitions \\
$w$ & 10 & loss weight: FIFA Club World Cup \\
 & 0.38 & goals of extra time, share of a regular match (from the data) \\
\tabend
\end{tabularx}\rowcolors{1}{}{}
\end{table}

The parameters are constrained to the values that have a meaning: the crowd and travel components of home advantage are non-negative, since playing at home is not expected to be a disadvantage, and the two components are separated only by the matches without spectators, so without the constraint one of them could become negative and be compensated by the other; promoted clubs do not become weaker ($\mu_{\text{lift}} \ge 0$); and the uncertainty added at the season transition, with the changes of the rosters, is not smaller than that of a rating period during the season ($\phi_{\text{season}} \ge \phi_{\text{period}} \ge 0$). The constraints mainly restrict the search to the meaningful region and speed up the estimation: the estimates of almost all leagues lie inside this region anyway (Table~\ref{tab:parameters}).

All parameters are estimated by minimizing the loss function described below with coordinate descent. The parameters of a single league change the loss only slightly, so a direct per-league search is slow and stops at plateaus. The estimation therefore goes from the most general parameters to the most specific ones: the global parameters, shifts of each parameter shared by all leagues, the level of each country (the initial ratings of its two divisions shifted together, since their gap is fixed by domestic cups and only international matches decide the level), the parameters of each league, and the parameters of extra time. The exact procedure is given in the code\footnote{\url{https://github.com/andreyshelopugin/SportsRatings/blob/main/soccer/club_ratings/train_glicko.py}, method \texttt{TrainGlicko.fit}.}. Finally, all initial ratings are shifted by a common constant so that the mean rating of the active clubs equals 1500, which does not change the predictions.

\label{subsec:country_shift}

The choice of the loss function follows from the purpose of the ratings. The goal is not only to predict individual matches well, but to estimate the current strength of clubs and leagues on a common scale, which is what the applications require (player evaluation, transfers, comparison of leagues). A model can predict domestic matches almost perfectly and still place the leagues at wrong levels relative to each other, so the loss is designed to emphasize the matches that carry information about the levels of leagues and the recent matches that carry information about the current strength.

The loss function is a weighted sum of the log losses of individual matches. Each match is weighted by the type of competition and by its age:
\begin{equation}
L = \sum_{t > t_0} \gamma^{\,T - t} \sum_{m \in \mathcal{M}_t} w_m \, \ell_m,
\qquad
w_m =
\begin{cases}
1, & \text{domestic match}, \\
4, & \text{continental club competition}, \\
10, & \text{FIFA Club World Cup},
\end{cases}
\label{eq:weighted_loss}
\end{equation}
where $\mathcal{M}_t$ is the set of matches of the rating period $t$, $T$ is the last rating period, $\ell_m$ is the log loss of match $m$ (including the extra time stage), and $\gamma$ is the discount factor.

The weights of international matches compensate for their small number. International matches make up less than 3\% of the dataset, but they are the only source of information about the relative levels of the countries: the levels of the leagues do not affect the predictions of domestic matches (Section~\ref{subsec:invariant}). Without the weights, the estimation is dominated by domestic matches, and a league whose clubs are systematically over- or underrated in international competitions costs only a small amount of the loss. The Club World Cup has the largest weight, since it is the main source of matches between confederations. The weights are used only for the estimation of the parameters: all reported test metrics are computed with equal weights, so that they are comparable with the metrics of the other models.

\begin{figure}[!htbp]
\centering
\includegraphics[width=\textwidth]{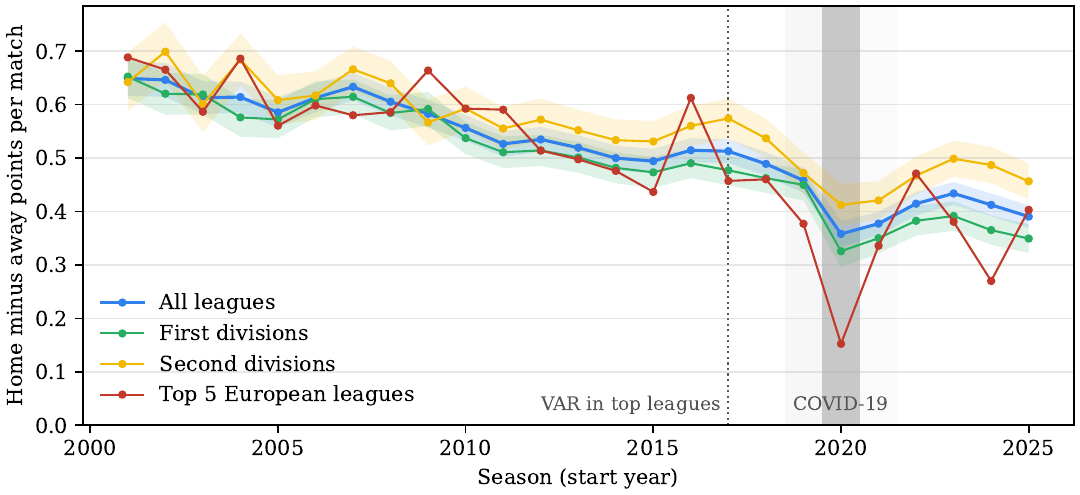}
\caption{Home advantage by season: the mean difference between the points of the home and the away team per match (3 for a win, 1 for a draw, after 90 minutes) in domestic league matches, with 95\% intervals, and separately for the first divisions of England, Spain, Germany, Italy and France (top 5; about 1,800 matches per season, the interval of about $\pm 0.12$ is not shown). The darkness of the band of a season is the share of its matches played during the COVID-19 pandemic (13\% in 2019, 93\% in 2020, 9\% in 2021). The dotted line marks the introduction of VAR in the top European leagues (2017/2018--2019/2020)}
\label{fig:home_advantage}
\end{figure}

The discount factor gives more weight to recent seasons, since the goal of the ratings is the current strength of the teams, and the parameters of the leagues change over time: home advantage, for example, has decreased over the last two decades (Fig.~\ref{fig:home_advantage}). A season contains four rating periods, and we use $\gamma = 0.95$ per rating period, which corresponds to a half-life of about 13.5 rating periods, or about 3.4 seasons.
The first $t_0 = 8$ rating periods (two seasons) are excluded from the loss, since the ratings are not yet established.

Home advantage decreased from about 0.65 points per match in the early 2000s to about 0.49 in 2018, fell to 0.36 in the 2020 season, when most matches were played without spectators (in the top 5 European leagues, where almost the whole season was played in empty stadiums, to 0.15), and has recovered only partly since then (0.39 in 2025). Home advantage in soccer has been documented for a long time~\cite{pollard1986home}, and part of it is attributed to the pressure of the crowd on the referees: referees favour the home team less when the crowd noise is absent~\cite{nevill2002influence} or the stadium is empty~\cite{pettersson2010behavior}, and the decline in 2017--2019 coincides with the introduction of the video assistant referee (VAR) in the top European leagues. However, the decline is visible in the second divisions as well, where VAR is mostly not used, so we do not attribute it to VAR. Whatever the causes, home advantage is not constant over time, which is another reason to discount older matches. A model with constant parameters estimated on the whole history would therefore overestimate home advantage in the current seasons, which is another reason to discount older matches.

Although it may be theoretically appealing to assign team-specific parameters—incorporating factors such as stadium capacity, geographic distance between cities, or other determinants of home advantage—such an approach substantially increases model complexity and typically results in overfitting.

The complete estimation pipeline of the clubs is shown in Figure~\ref{fig:pipeline_clubs} in Appendix~\ref{app:pipeline}.

\subsection{League Ratings}\label{subsec:league_ratings}

The rating of a league is computed from the current ratings of its clubs as a weighted mean of the ratings of the eight strongest clubs:
\begin{equation}
R_{\text{league}} = \frac{\sum_{k=1}^{8} w_k \, \mu_{(k)}}{\sum_{k=1}^{8} w_k}, \qquad w_k = 1.1 - 0.1\,k,
\label{eq:league_rating}
\end{equation}
where $\mu_{(k)}$ is the rating of the club ranked $k$-th in the league, so the weights decrease from 1 for the strongest club to 0.3 for the eighth. The number eight is the size of the smallest leagues in the dataset, so every league is measured by the same number of clubs. The decreasing weights reflect the fact that the level of a league is perceived mainly through its strongest clubs, which represent the league in the continental competitions. At the same time, unlike the UEFA coefficient, the rating takes into account the whole upper half of a typical league, not only the clubs that play in international competitions.

The model places all leagues on one scale, but the precision of the level of a league depends on how strongly it is connected to the other leagues. A league that regularly plays in continental competitions is anchored by hundreds of matches, while the level of an isolated league is inferred mainly from its initialization and from a few matches of its clubs. Even a few matches against other leagues or a few promoted and relegated teams link a league to the common scale, which is the purpose of the model. A league without any of these links over the last three seasons, however, keeps the level of its initialization. Such leagues (6 of 250: the first divisions of Bermuda, Macao, Palestine, Sri Lanka and Yemen, and the second division of Mexico, which has no promotion to the first division) are rated by the model, but are not shown in Table~\ref{tab:league_ratings}. The same holds for a country as a whole: the two Russian divisions are linked to each other, but their clubs have played almost no international matches since 2022, so the level of the country is not identified and the Russian leagues are not shown either. The weighted top-8 is not the only reasonable definition of the strength of a league, and the best definition depends on the task. For the assessment of a transfer from a top club, the rating of the strongest clubs is more relevant; for the evaluation of a typical player of the league, the mean or the median over all clubs may be more appropriate. Since the ratings of all clubs are available, any such aggregate can be computed directly; the tables of the repository also include the means of the top 5, top 7 and top 10 clubs and the ratings of the clubs by position.

The ratings of the leagues and clubs after the 2025/2026 season are given in Tables~\ref{tab:league_ratings} and~\ref{tab:club_ratings}, with the strongest club, the median club and the standard deviation (SD) over all clubs. The aggregates can rank leagues differently. The Saudi Pro League is ahead of the Brazilian Serie A by the weighted top 8 (2093 against 2067) and by the mean of the top 5 clubs (2141 against 2083), since its strongest clubs are stronger. The Brazilian league, however, is much more even (standard deviation 89 against 189), and its median club is stronger by 120 points (1940 against 1820), so a typical match in Brazil is played at a higher level.

\section{Experimental Results}

All results are computed on the held-out seasons according to the protocol of Section~\ref{subsec:protocol}.

\subsection{Metrics}\label{subsec:metrics}

Let $p_{m,k}$ be the predicted probability of the outcome $k \in \{\text{H}, \text{D}, \text{A}\}$ of match $m$, and $y_{m,k} = 1$ if the outcome occurred and 0 otherwise. The main metric is the outcome log loss, the mean of the binary log losses of the three outcomes:
\begin{equation}
\mathrm{LL} = -\frac{1}{3N} \sum_{m=1}^{N} \sum_{k} \Big( y_{m,k} \log p_{m,k} + (1 - y_{m,k}) \log (1 - p_{m,k}) \Big),
\label{eq:outcome_log_loss}
\end{equation}
which is also the loss used in the estimation of the Glicko parameters, so the values are directly comparable across all models. We also report the multiclass Brier score
\begin{equation}
\mathrm{BS} = \frac{1}{N} \sum_{m=1}^{N} \sum_{k} (p_{m,k} - y_{m,k})^2
\label{eq:brier}
\end{equation}
and the ranked probability score (RPS), which takes into account the order of the outcomes (a home win is closer to a draw than to an away win)~\cite{constantinou2012solving}:
\begin{equation}
\mathrm{RPS} = \frac{1}{2N} \sum_{m=1}^{N} \sum_{r=1}^{2} \Big( \sum_{k \le r} (p_{m,k} - y_{m,k}) \Big)^2 ,
\label{eq:rps}
\end{equation}
where the outcomes are ordered as H, D, A. Lower values are better for all three metrics.

\subsection{Comparison of Score Matrix Models}\label{subsec:score_models}

The score-matrix models serve as baselines that share the goal model and differ only in how the goal expectations are converted into outcome probabilities. Since all of them use the same expected goals, the comparison isolates this last step and can be viewed as an ablation of the structure of the score distribution.

The parameters of the score-matrix models ($\rho$ of Dixon--Coles and the nine parameters of the Markov model) are estimated with the expected goals of LightGBM as inputs. They are fitted on a held-out 10\% of the training seasons (90,979 matches), on which LightGBM was not trained (these matches were used only for early stopping). As in stacking, the second-level model is fitted on predictions for new data, as it is used in practice, rather than on the training matches of the first-level model.

We compare three such models: the Skellam probabilities (independent Poisson goals), the Dixon--Coles matrix and the Markov score matrix (Table~\ref{tab:score_models}).

\begin{table}[!htbp]
\caption{Score matrix models on the test seasons (90 minutes): outcome log loss of clubs (311,166 matches), national teams (3,897) and all matches, Brier score and RPS of all matches. All models use the same goal expectations of LightGBM. Bold: the best value in a column}
\label{tab:score_models}
\centering\sffamily\renewcommand{\arraystretch}{1.2}\tabtop
\begin{tabularx}{\textwidth}{X>{\raggedleft\arraybackslash}p{1.75cm}>{\raggedleft\arraybackslash}p{1.75cm}>{\raggedleft\arraybackslash}p{1.75cm}>{\raggedleft\arraybackslash}p{1.75cm}>{\raggedleft\arraybackslash}p{1.75cm}}
\rowcolor{M}\hd{Model} & \hd{Clubs} & \hd{National} & \hd{All} & \hd{Brier} & \hd{RPS} \\
Skellam (independent Poisson) & 0.583899 & 0.518382 & 0.583089 & 0.596645 & 0.203256 \\
Dixon--Coles & 0.583638 & 0.518363 & 0.582831 & 0.596330 & 0.203203 \\
Markov & \textbf{0.583569} & \textbf{0.517802} & \textbf{0.582756} & \textbf{0.596266} & \textbf{0.203189} \\
\tabend
\end{tabularx}\rowcolors{1}{}{}
\end{table}

The independent Poisson score matrix (Eq.~\eqref{eq:independent_poisson_matrix}) gives the same probabilities as the Skellam distribution up to the truncation of the matrix, so it is not reported separately.

The Dixon--Coles and Markov models achieve slightly lower loss than the independent Poisson model on club matches: the Dixon--Coles correction reduces the loss by 0.00026 and the Markov model by 0.00033. For national teams, the differences are of the same size or smaller than the noise of a sample of 3,897 matches. Overall, the structure of the score distribution matters much less than the goal expectations themselves.

\subsection{Comparison with the Baselines}\label{subsec:main_results}

As a baseline rating model we use the classical Elo rating~\cite{elo1967proposed} with home advantage and the draw probabilities of Davidson~\cite{davidson1970extending}:
\begin{equation}
P(\text{H}) = \frac{q_i}{D}, \quad P(\text{A}) = \frac{q_j}{D}, \quad P(\text{D}) = \frac{\nu \sqrt{q_i q_j}}{D}, \qquad
q_i = 10^{(r_i + h x_{ij}) / 400}, \quad q_j = 10^{r_j / 400},
\label{eq:elo}
\end{equation}
where $D$ normalizes the probabilities, and the ratings are updated after every match by $K$ times the difference between the actual and the expected score. As usual for Elo, all clubs start from the same rating. The parameters are estimated by a grid search on the training seasons ($K$ from 10 to 50, $h$ from 40 to 160 and $\nu$ from 0.6 to 1.1), which gives $K = 35$, $h = 100$ and $\nu = 0.8$; none of the estimates lies at the boundary of the grid.

Table~\ref{tab:main_results} compares all models on the club matches of the test seasons, separately for domestic and international matches. For each model, the last column gives the difference in log loss between the modified Glicko model and the model, with a 95\% interval from a paired bootstrap over matches (1,000 resamples); a negative difference means that Glicko is better.

\begin{table}[!htbp]
\caption{Club matches of the test seasons (90 minutes): outcome log loss, Brier score and RPS. $\Delta$: log loss of the modified Glicko model minus log loss of the model, with a 95\% paired bootstrap interval (negative: Glicko is better). Glicko + Markov is the average of the probabilities of the two models. Bold: the best model in a group; in international matches, the modified Glicko model and Glicko + Markov do not differ significantly, and both are bold}
\label{tab:main_results}
\centering\sffamily\renewcommand{\arraystretch}{1.2}\tabtop
\begin{tabularx}{\textwidth}{X>{\raggedleft\arraybackslash}p{1.9cm}>{\raggedleft\arraybackslash}p{1.9cm}>{\raggedleft\arraybackslash}p{1.9cm}>{\raggedleft\arraybackslash}p{4.6cm}}
\rowcolor{M}\hd{Model} & \hd{Log loss} & \hd{Brier} & \hd{RPS} & \hd{$\Delta$ [95\% interval]} \\
\sect{5}{All matches (311,152)}
Elo & 0.591105 & 0.606114 & 0.207224 & \textminus 0.0075 [\textminus 0.0079, \textminus 0.0070] \\
Skellam & 0.583895 & 0.597624 & 0.203627 & \textminus 0.0002 [\textminus 0.0007, 0.0002] \\
Dixon--Coles & 0.583634 & 0.597304 & 0.203573 & 0.0000 [\textminus 0.0004, 0.0005] \\
Markov & 0.583565 & 0.597245 & 0.203561 & 0.0001 [\textminus 0.0004, 0.0005] \\
Modified Glicko & 0.583646 & 0.596989 & 0.203507 & -- \\
Glicko + Markov & \textbf{0.580274} & \textbf{0.593186} & \textbf{0.201684} & 0.0034 [0.0031, 0.0036] \\
\sect{5}{Domestic matches (301,671)}
Elo & 0.590887 & 0.605871 & 0.206943 & \textminus 0.0063 [\textminus 0.0067, \textminus 0.0059] \\
Skellam & 0.584536 & 0.598446 & 0.203853 & 0.0000 [\textminus 0.0004, 0.0004] \\
Dixon--Coles & 0.584256 & 0.598101 & 0.203795 & 0.0003 [\textminus 0.0001, 0.0007] \\
Markov & 0.584202 & 0.598058 & 0.203791 & 0.0003 [\textminus 0.0001, 0.0008] \\
Modified Glicko & 0.584546 & 0.598129 & 0.203889 & -- \\
Glicko + Markov & \textbf{0.581084} & \textbf{0.594223} & \textbf{0.202018} & 0.0035 [0.0032, 0.0037] \\
\sect{5}{International matches (9,481)}
Elo & 0.598042 & 0.613843 & 0.216154 & \textminus 0.0430 [\textminus 0.0469, \textminus 0.0392] \\
Skellam & 0.563512 & 0.571475 & 0.196416 & \textminus 0.0085 [\textminus 0.0114, \textminus 0.0056] \\
Dixon--Coles & 0.563853 & 0.571947 & 0.196495 & \textminus 0.0088 [\textminus 0.0119, \textminus 0.0060] \\
Markov & 0.563290 & 0.571379 & 0.196254 & \textminus 0.0083 [\textminus 0.0113, \textminus 0.0054] \\
Modified Glicko & \textbf{0.555016} & \textbf{0.560713} & \textbf{0.191361} & -- \\
Glicko + Markov & \textbf{0.554494} & \textbf{0.560188} & \textbf{0.191027} & 0.0005 [\textminus 0.0010, 0.0020] \\
\tabend
\end{tabularx}\rowcolors{1}{}{}
\end{table}

The modified Glicko model is clearly better than Elo in all groups of matches. Over all matches and over domestic matches, it is as accurate as the score-matrix models: the intervals of the differences in log loss include zero, and by the Brier score and the RPS it is the best single model over all matches.
Both approaches are built on the same matches, so the comparable accuracy is expected. The difference is in what else they give: the rating model also gives a single interpretable number for every club, comparable across leagues and seasons, together with its uncertainty.

In international matches, however, the modified Glicko model is substantially more accurate than all baselines: its log loss is lower by about 0.009 than that of the score-matrix models and by 0.043 than that of Elo, with intervals far from zero. International matches are the matches in which the levels of different leagues meet, and they are exactly the matches that the league-aware components of the model are designed for. The goal model describes the scoring of a team by its recent matches, which are mostly played within its own league, so it does not know how strong the league is; Elo, with the same initial rating for all clubs, does not know it either. The ablation below attributes the advantage to the league-aware initialization.

The test metrics are computed without the weights of international matches and without the discounting used in the estimation of the Glicko parameters (Section~\ref{subsec:parameter_estimation}), so the Glicko model is evaluated by a criterion it does not optimize, while the score-matrix models are fitted on the same criterion they are evaluated by. The comparison is therefore conservative for the rating model.

Finally, the average of the probabilities of the modified Glicko and the Markov models is the best model in all groups, with a large and significant gain over each of them in domestic matches. The two approaches use different information: the rating summarizes the whole history of results and the levels of the leagues, while the goal model describes the recent scoring of the teams. In international matches, the average is as accurate as Glicko alone, since the information about the levels of the leagues comes from the rating.

\subsection{Ablation}\label{subsec:ablation}

To measure the contribution of the components of the model, we switch them off one at a time and compute the test loss (Table~\ref{tab:ablation}). In the version without refit, the other parameters are kept as estimated; in the version with refit, the global parameters and the parameters shared by the leagues are re-estimated on the training seasons without the component, so that the other parameters can compensate for it; the parameters of the individual leagues are kept as estimated.

\begin{table}[!htbp]
\caption{Ablation: increase of the test log loss of the modified Glicko model (club matches, 90 minutes) when one component is switched off. Without refit: the other parameters are fixed; with refit: the global and shared parameters are re-estimated without the component}
\label{tab:ablation}
\centering\sffamily\renewcommand{\arraystretch}{1.2}\tabtop
\begin{tabularx}{\textwidth}{X>{\raggedleft\arraybackslash}p{1.55cm}>{\raggedleft\arraybackslash}p{1.55cm}>{\raggedleft\arraybackslash}p{1.55cm}>{\raggedleft\arraybackslash}p{1.55cm}>{\raggedleft\arraybackslash}p{1.55cm}>{\raggedleft\arraybackslash}p{1.55cm}}
\rowcolor{M} & \multicolumn{3}{c}{\hd{Without refit}} & \multicolumn{3}{c}{\hd{With refit}} \\
\rowcolor{M}\hd{Switched off} & \hd{All} & \hd{Domestic} & \hd{Intern.} & \hd{All} & \hd{Domestic} & \hd{Intern.} \\
Home advantage & 0.0097 & 0.0094 & 0.0189 & 0.0096 & 0.0094 & 0.0181 \\
League-aware initialization & 0.0018 & 0.0005 & \textbf{0.0443} & 0.0019 & 0.0006 & \textbf{0.0433} \\
Scoring dependence ($\alpha_2$, $\beta_2$) & 0.0021 & 0.0021 & 0.0018 & 0.0019 & 0.0019 & 0.0018 \\
Season growth of $\phi$ & 0.0013 & 0.0014 & \textminus 0.0008 & 0.0013 & 0.0014 & \textminus 0.0010 \\
Newcomer offset $\mu_{\text{new}}$ & 0.0003 & 0.0003 & 0.0001 & 0.0003 & 0.0003 & \textminus 0.0002 \\
Promotion/relegation $\mu_{\text{lift}}$ & 0.0000 & 0.0000 & 0.0000 & 0.0001 & 0.0001 & \textminus 0.0001 \\
\tabend
\end{tabularx}\rowcolors{1}{}{}
\end{table}

The league-aware initialization hardly matters for domestic matches, where the level of the league cancels out, but without it the loss of international matches increases by 0.044, about the same as the gap between Glicko and Elo in these matches. The advantage of the proposed model over Elo in international matches is therefore due to the initialization of the clubs at the level of their league. Home advantage is the most important component overall. The dependence of home advantage and of draws on the expected goals reduces the loss by about 0.002 in all groups; since the strength of the teams is already in the model, this is information that the ratings do not contain, which answers the question whether the expected goals are redundant given the ratings (stronger teams also score more). The season growth of the rating deviation helps in domestic matches and slightly hurts in international ones. The newcomer offset and the promotion/relegation adjustment hardly change the loss: their role is not the accuracy of individual matches but the stability of the levels of the leagues over time (Section~\ref{subsec:invariant}), which is checked by the calibration below. The refit changes the effects only slightly: the other parameters compensate for a removed component only to a small extent, so the contributions without refit are not exaggerated.

\subsection{Calibration of League Levels}\label{subsec:calibration}

Log loss measures the quality of predictions over all matches, but it says little about whether the levels of different leagues are consistent with each other. Most matches are played within a single league, so a league that is rated too high or too low as a whole may cost only a small amount of log loss.
We therefore complement log loss with a calibration check.

For every match, the expected score of a team is computed from the pre-match outcome probabilities of the model,
\begin{equation}
E[\text{score}] = P(\text{win}) + \tfrac{1}{2} P(\text{draw}),
\end{equation}
and compared with the actual score (1, 0.5 or 0). The difference between the actual and the expected score is averaged over groups of matches; if the model is calibrated for a group, the average difference is close to zero. The standard deviation of the score in a single match is about 0.44, so for a group of $n$ matches the noise level of the average difference is approximately $0.44/\sqrt{n}$, for example 0.044 for 100 matches.

We consider two types of groups. The first are the teams in their first season after a league transition (promoted, relegated, new in the league, or remaining in the same league), in domestic matches since 2010. These groups check the rules of Section~\ref{subsec:invariant}. The second are the leagues in international club matches: if a league is underrated, its clubs score more in international competitions than the model expects. This is the most direct test of the levels of the leagues. Since international matches also enter the loss function with large weights, we perform this test on the test seasons: the parameters are estimated on the seasons before 2021, and the predictions are the sequential predictions of Section~\ref{subsec:protocol}, so the matches being checked were not used in any way.

The results are given in Table~\ref{tab:calibration}. Teams that remain in their league, promoted teams and newcomers are well calibrated. Relegated teams score slightly less than expected in their first season in the lower league ($-0.021$ per match).

In the international matches of the test seasons, the mean absolute difference over the 142 leagues with at least 20 matches is 0.042, while the noise alone gives about 0.040 for a perfectly calibrated model (the mean absolute value of a normal variable is $0.8$ of its standard deviation). The difference exceeds two noise levels in 4.2\% of the leagues, as expected by chance. The leagues with the most international matches are calibrated within about one noise level (Brazil $+0.002$, Argentina $+0.018$, England $+0.022$, Spain $-0.024$, Germany $+0.008$). Some of the largest deviations follow from changes of the level of a league. Gradual changes of the level of a league are followed by the ratings, but sudden changes are followed with a delay, since the level of a league is corrected only by its international matches. The Saudi Pro League ($+0.051$) has become much stronger since 2023, when its clubs started to sign star players, and the Ukrainian league ($-0.082$) has become weaker since 2022, when many players left the country and the clubs started to play their home matches abroad. The deviations of the Asian leagues (Japan $+0.111$, about 3 noise levels, and Iran $-0.050$) are not related to such changes and are left for future work (Section~\ref{discussion}).

\begin{table}[!htbp]
\caption{Calibration of the modified Glicko model: actual minus expected score per match (win 1, draw 0.5, loss 0). Transitions: domestic matches since 2010. Leagues: international club matches of the test seasons, predicted with the parameters estimated before 2021: the four leagues with the most matches and the leagues with the largest deviations, and the mean over all 143 leagues with at least 20 matches (the full table is in the repository). The noise level is $0.44/\sqrt{n}$. I: first division}
\label{tab:calibration}
\centering\sffamily\renewcommand{\arraystretch}{1.2}\tabtop
\begin{tabularx}{\textwidth}{X>{\raggedleft\arraybackslash}p{2.6cm}>{\raggedleft\arraybackslash}p{3.4cm}>{\raggedleft\arraybackslash}p{2cm}}
\rowcolor{M}\hd{Group} & \hd{Matches} & \hd{Actual $-$ Expected} & \hd{Noise} \\
\sect{4}{First season after a transition}
Same league & 1,061,062 & 0.001 & 0.000 \\
Promoted & 71,462 & 0.000 & 0.002 \\
Relegated & 59,648 & \textminus 0.021 & 0.002 \\
New in the league & 176,116 & 0.006 & 0.001 \\
\sect{4}{International matches of the test seasons, selected leagues}
Brazil I & 853 & 0.002 & 0.015 \\
Argentina I & 603 & 0.018 & 0.018 \\
England I & 450 & 0.022 & 0.021 \\
Spain I & 411 & \textminus 0.024 & 0.022 \\
Japan I & 181 & 0.111 & 0.033 \\
Saudi Arabia I & 196 & 0.051 & 0.031 \\
Iran I & 120 & \textminus 0.050 & 0.040 \\
Ukraine I & 160 & \textminus 0.082 & 0.035 \\
\totalrule\bfseries Mean $|$difference$|$, 142 leagues & & \bfseries 0.042 & \bfseries 0.040 \\
\tabend
\end{tabularx}\rowcolors{1}{}{}
\end{table}

\subsection{Estimated Parameters}\label{subsec:parameters}

The estimates of all parameters are given in Table~\ref{tab:parameters}. They are stable: the estimates on the training seasons and on all seasons are close (for example, the mean crowd component 0.119 and 0.101, the draw baseline $-0.265$ and $-0.245$, $\mu_{\text{lift}}$ 29 and 38, and the initial ratings of the leagues differ by 12 points on average).

The league parameters are interpretable. The median home advantage $\alpha_{\mathrm{crowd}} + \alpha_{\mathrm{logistic}}$ is 0.22 in the first divisions and 0.23 in the second divisions. Among the leagues with at least 1,000 matches, it is the largest in Nigeria (1.00 in the first division and 0.87 in the second), followed by Bosnia and Herzegovina, Algeria, Guatemala and Ghana (about 0.6). Draws are the most frequent, relative to the strength and the expected goals of the teams, in El Salvador, the United Arab Emirates, Iran and the Italian Serie B, and the least frequent in the second divisions of Bosnia and Herzegovina and Albania and in Kazakhstan. The parameters of small leagues stay close to their starting values, since a single league changes the loss only slightly, so they should not be interpreted.

The extra time parameters show what changes in a short game after a level score. The difference in strength matters less ($\kappa = 0.76$). This is partly a selection effect rather than a property of extra time: extra time follows a draw, which is evidence that the teams were closer on the day than their ratings suggest. The home advantage is about 60\% of that in regular time ($s_{\mathrm{ET}} = 0.58$), and the large draw baseline reflects the fact that 40\% of extra time periods end without a goal. The mechanism of draws is the same as in regular time: per unit of the expected goals of 90 minutes, the effect of scoring on draws is $-1.63 \cdot 0.38 = -0.62$ in extra time and $-0.50$ on average in regular time, and the effect of strength is the same ($\beta_1^{\mathrm{ET}} = 1.00$ and $\beta_1 = 1.01$).

\subsection{Interpretation of Ratings}\label{subsec:interpretation}

The ratings have a direct probabilistic interpretation: Table~\ref{tab:interpretation} gives the outcome probabilities for a given rating difference.

\begin{table}[!htbp]
\caption{Outcome probabilities (\%) by the rating difference between the stronger and the weaker team, at a neutral venue and at the home stadium of the stronger team (parameters of the English Premier League, a match with the average expected number of goals)}
\label{tab:interpretation}
\centering\sffamily\renewcommand{\arraystretch}{1.2}\tabtop
\begin{tabularx}{\textwidth}{>{\raggedleft\arraybackslash}p{2.6cm}>{\raggedleft\arraybackslash}X>{\raggedleft\arraybackslash}X>{\raggedleft\arraybackslash}X>{\raggedleft\arraybackslash}X>{\raggedleft\arraybackslash}X>{\raggedleft\arraybackslash}X}
\rowcolor{M}\hd{Rating} & \multicolumn{3}{c}{\hd{Neutral venue}} & \multicolumn{3}{c}{\hd{Home of the stronger team}} \\
\rowcolor{M}\hd{difference} & \hd{Win} & \hd{Draw} & \hd{Loss} & \hd{Win} & \hd{Draw} & \hd{Loss} \\
\sect{7}{Regular time}
0 & 36.5 & 27.1 & 36.5 & 44.4 & 26.6 & 29.0 \\
10 & 37.5 & 27.1 & 35.4 & 45.5 & 26.5 & 28.0 \\
50 & 41.8 & 26.9 & 31.3 & 49.8 & 25.9 & 24.4 \\
100 & 47.2 & 26.3 & 26.5 & 55.0 & 24.8 & 20.2 \\
200 & 57.7 & 24.1 & 18.2 & 64.8 & 21.9 & 13.4 \\
300 & 67.1 & 21.0 & 11.9 & 73.0 & 18.5 & 8.5 \\
500 & 81.2 & 14.3 & 4.6 & 84.8 & 12.1 & 3.1 \\
\sect{7}{Extra time}
0 & 21.1 & 57.9 & 21.1 & 23.7 & 57.7 & 18.6 \\
10 & 21.5 & 57.9 & 20.6 & 24.2 & 57.6 & 18.1 \\
50 & 23.4 & 57.7 & 18.8 & 26.3 & 57.2 & 16.5 \\
100 & 26.0 & 57.3 & 16.7 & 29.0 & 56.5 & 14.6 \\
200 & 31.4 & 55.6 & 13.0 & 34.6 & 54.2 & 11.2 \\
300 & 37.2 & 52.9 & 10.0 & 40.5 & 51.0 & 8.5 \\
500 & 49.3 & 45.2 & 5.5 & 52.7 & 42.7 & 4.6 \\
\tabend
\end{tabularx}\rowcolors{1}{}{}
\end{table}

\subsection{National Teams}\label{subsec:national_results}

The national teams are rated by the same model as the clubs, with the confederations in place of the leagues: a team enters with the initial rating of its confederation, and the levels of the confederations are estimated from the matches between them, mainly at the World Cup. The host of a final tournament is treated as the usual home team, and the other matches of a final tournament are played at a neutral venue. The home advantage parameters are global (Section~\ref{subsec:negative}). In the loss, the matches of the World Cup final tournaments have weight 3, the matches of the other final tournaments weight 2, and the qualifiers and the Nations Leagues weight 1. As for the clubs, the rating periods are the calendar quarters; since national teams do not play in every quarter, the rating deviation grows in proportion to the time between two periods with matches, and there is no additional growth at season transitions. Extra time is modelled as for the clubs (Section~\ref{subsec:extra_time}). The pipeline is shown in Figure~\ref{fig:pipeline_national}.

The parameters are estimated on the matches before 2021 and evaluated on the matches since 2021 (the qualifiers for the 2022 World Cup and later), with sequential predictions as for the clubs. Elo is estimated in the same way as for the clubs (Section~\ref{subsec:main_results}); the grid search gives $K = 80$, $h = 100$ and $\nu = 0.7$, so the ratings of national teams, which play only a few matches per year, change faster than those of the clubs ($K = 35$). Table~\ref{tab:national_results} compares the models on the 3,637 matches that are in the test sets of all models. The goal model of the baselines is split by season, so its training data include a few final tournaments played in 2021 (for example, Euro 2020), which, if anything, favours the baselines.

\begin{table}[!htbp]
\caption{National teams, test matches since 2021 (90 minutes): outcome log loss, Brier score and RPS. $\Delta$: log loss of the modified Glicko model minus log loss of the model, with a 95\% paired bootstrap interval (negative: Glicko is better). Final tournaments: the World Cup and the continental championships}
\label{tab:national_results}
\centering\sffamily\renewcommand{\arraystretch}{1.2}\tabtop
\begin{tabularx}{\textwidth}{X>{\raggedleft\arraybackslash}p{1.9cm}>{\raggedleft\arraybackslash}p{1.9cm}>{\raggedleft\arraybackslash}p{1.9cm}>{\raggedleft\arraybackslash}p{4.6cm}}
\rowcolor{M}\hd{Model} & \hd{Log loss} & \hd{Brier} & \hd{RPS} & \hd{$\Delta$ [95\% interval]} \\
\sect{5}{All matches (3,637)}
Elo & 0.510621 & 0.509617 & 0.168752 & \textminus 0.0106 [\textminus 0.0145, \textminus 0.0063] \\
Skellam & 0.519458 & 0.519064 & 0.174164 & \textminus 0.0194 [\textminus 0.0251, \textminus 0.0134] \\
Dixon--Coles & 0.519460 & 0.519140 & 0.174176 & \textminus 0.0194 [\textminus 0.0252, \textminus 0.0134] \\
Markov & 0.518880 & 0.518611 & 0.173952 & \textminus 0.0188 [\textminus 0.0248, \textminus 0.0127] \\
Modified Glicko & \textbf{0.500060} & \textbf{0.496356} & \textbf{0.163617} & -- \\
Glicko + Markov & 0.501617 & 0.497917 & 0.164305 & \textminus 0.0016 [\textminus 0.0046, 0.0015] \\
\sect{5}{Final tournaments (592)}
Elo & 0.581951 & 0.594969 & 0.191977 & \textminus 0.0373 [\textminus 0.0530, \textminus 0.0202] \\
Skellam & 0.581367 & 0.595744 & 0.195374 & \textminus 0.0367 [\textminus 0.0542, \textminus 0.0189] \\
Dixon--Coles & 0.580276 & 0.594539 & 0.195173 & \textminus 0.0356 [\textminus 0.0532, \textminus 0.0176] \\
Markov & 0.580249 & 0.594666 & 0.195150 & \textminus 0.0356 [\textminus 0.0534, \textminus 0.0173] \\
Modified Glicko & \textbf{0.544675} & \textbf{0.547217} & \textbf{0.172870} & -- \\
Glicko + Markov & 0.551509 & 0.556765 & 0.177274 & \textminus 0.0068 [\textminus 0.0157, 0.0026] \\
\tabend
\end{tabularx}\rowcolors{1}{}{}
\end{table}

For national teams, the modified Glicko model is significantly more accurate than all baselines, by about 0.019 in log loss than the score-matrix models and by 0.011 than Elo. Unlike for the clubs, the score-matrix models are less accurate than Elo, and the average with the Markov model does not improve on Glicko alone. The gap is the largest in the final tournaments (0.036), where teams of different confederations meet; the ablation below attributes it to the levels of the confederations, as the advantage in international club matches is attributed to the levels of the leagues.

The ablation confirms this. Without the levels of the confederations (all teams start from the same rating), the test loss increases by 0.007 over all matches and by 0.028 in the final tournaments; with a refit of the other parameters, by 0.004 and 0.021, so the other parameters cannot compensate for the levels. Without home advantage, the loss increases by 0.003 (0.003 with a refit), mostly in the qualifiers, where every match has a host. The dependence on the expected goals adds 0.004 without a refit but almost nothing with a refit (0.0004), since the draw parameters take over its role.

The estimated initial levels of the confederations relative to Europe are $+415$ for South America, $-295$ for Africa, $-653$ for Asia, $-700$ for North and Central America and $-1153$ for Oceania. The ratings after the 2026 World Cup (Table~\ref{tab:national_ratings}) are led by Spain (2421), Argentina (2345), France (2321) and England (2274), the same four teams as at the top of the FIFA ranking. The largest differences from the FIFA ranking concern the hosts of the 2026 World Cup and the Asian teams, which are lower in our ratings (Mexico 19th, Japan 27th, the USA 34th): for the hosts, one reason is that a large part of their recent results was obtained at home, and the model, unlike the FIFA ranking, discounts the home advantage. Another difference is that the FIFA ranking also counts friendly matches, which are not in our data.

\subsection{Application: League Ratings in Player Performance Prediction}
Consider, for instance, our objective to predict various player metrics for the upcoming season, such as the number of goals a specific player will score. To make accurate predictions, we can take into account factors like player age, position, previous performance, and the number of goals scored in the previous season, along with other relevant features. However, a significant challenge arises when a player undergoes transfers, moving to a different club or league during their career. How can we effectively inform the model about such transitions?

One possible solution involves introducing a categorical feature, such as "player transfers from the Denmark Superliga to France Ligue 1." Nonetheless, this approach encounters the curse of dimensionality—there are relatively few transfers between these two leagues, making it challenging for the model to train effectively. A more viable approach, which aligns with our method, is to generate numeric features. For example, we can create a feature that quantifies the difference in ratings between the Denmark Superliga and France Ligue 1. This enables the model to account for league transitions more effectively. Features of this kind were used to forecast the impact of a transfer, with the strength of the clubs and leagues measured by Elo ratings~\cite{dinsdale2022transfer}, and to predict the expected possession value of players in the next season~\cite{shelopugin2026expected}.

The ratings can also be used to evaluate the difficulty of a schedule. The expected number of points of a team over a sequence of matches, computed from the ratings of the opponents and the venues, measures how demanding a period of the season is. Combined with the dates of the matches, this allows clubs to plan the rotation of players (load management): for example, to rest key players in matches against weaker opponents between two difficult matches, or to estimate the cost of a congested period with several strong opponents in a row.

Additionally, we can enhance the precision of player performance analysis, particularly within the context of a single team. For instance, we can calculate the average rating of the opposing teams faced by each player. This fine-grained analysis helps differentiate player performances, as several players may exhibit similar statistics, but the strength of the opponent teams they encounter can vary significantly.

\subsection{Negative Results}\label{subsec:negative}

Several variants of the model were tested and did not improve it, so they are not used:
\begin{itemize}[label={--}, nosep, leftmargin=*]
\item separate parameters of promotion and relegation for every league (as in the earlier version of the model~\cite{shelopugin2026constructing}) inflated the levels of some leagues and were replaced by the single parameter $\mu_{\text{lift}}$ (Section~\ref{subsec:invariant});
\item a stricter version of the newcomer rule, in which the continuing teams of a league pay the offset of the newcomers, made the calibration of the newcomers, the relegated teams and the international matches worse (Section~\ref{subsec:invariant});
\item $\phi_{\text{season}}$ estimated for every league: the loss was almost flat in this parameter, and the only systematic difference was between the first and the second divisions, so two global values are used;
\item the offset of the newcomers $\mu_{\text{new}}$ estimated for every league: a single league changes the loss by less than one unit, so the values are not identifiable;
\item the volatility of Glicko-2, an individual speed of the rating of every team, changed the loss only marginally, for the clubs as well as for the national teams;
\item for the national teams, rating periods that follow their calendar (every international window and every match day of a final tournament as a separate period) gave the same test loss as the calendar quarters;
\item for the national teams, parameters for every country (home advantage and initial rating, as offsets to the global and confederation values) overfitted the training matches and did not reduce the test loss;
\item for the national teams, a larger home advantage of the host of a final tournament (about 2.5 times the usual one) made the test loss worse, so the host is treated as a usual home team;
\item for the national teams, weights of the rating update by the importance of the match (as the K-factor of the World Football Elo ratings) and a lower weight of the matches in which a team has already qualified (dead rubbers) did not reduce the test loss;
\item for the national teams, scaling the rating change by the goal difference did not improve on a uniformly larger rating change.
\end{itemize}

Some choices were not tuned and are left for future work. For the clubs, only the calendar quarters were used as rating periods. The weights of international matches in the loss (4 for the continental competitions and 10 for the Club World Cup) and the discount factor $\gamma = 0.95$ were set by judgement rather than estimated (Table~\ref{tab:parameters}), and the parameters are estimated once rather than in an expanding window (Section~\ref{discussion}).

\section{Discussion}\label{discussion}

The primary limitation of the proposed framework is its reliance exclusively on match outcomes. As a result, the model implicitly assumes that team strength evolves smoothly over time. In practice, however, roster composition can change substantially during the season, particularly in second-tier leagues where transfers and financial instability are more common. Such structural breaks are only partially captured through seasonal and mid-season adjustments.

The dataset spans six confederations, providing broad global coverage. However, direct intercontinental competition remains relatively rare. Until 2025, matches between clubs from different confederations were limited to a few matches of the FIFA Club World Cup per year. The expanded Club World Cup of 32 clubs (since 2025) and, within Europe, the Conference League (since 2021/2022) substantially increase the number of matches that connect leagues of different levels and continents, and we expect the cross-confederation levels to become more reliable as these tournaments accumulate. At present, the sparse cross-confederation interaction still reduces the amount of information available to anchor ratings across regions and introduces additional uncertainty when comparing clubs from different continents; for national teams, the same role is played by the World Cup.

At the same time, the reliance on match outcomes makes the model undemanding: it requires only the results of matches and a small amount of metadata (the venue, the type of competition, the league of each club), which are available for almost every professional league in the world. This is what makes a worldwide coverage possible, since richer data such as expected goals or player-level data exist only for a small number of top leagues. Where such data are available, the model is easy to extend. In an earlier experiment, for example, we replaced the actual outcome of a match with the outcome by expected goals (xG)~\cite{pollard1997measuring}, that is, the win was credited to a team whose xG exceeded the xG of the opponent by more than 0.5, and the match was counted as a draw otherwise. More generally, any signal that can be expressed as an outcome or as an adjustment of the scoring intensity can enter the model without changing its structure.

The framework is not specific to soccer. Its components (strength-dependent draws, home advantage, a scoring intensity from a goal or points model, league-aware initialization and transitions between divisions) apply to any team sport with a league structure, promotion and relegation or international club competitions. Basketball, handball, volleyball and ice hockey have national leagues of very different levels connected by continental competitions, and the same problem of cross-league comparison arises there. Sports without draws (e.g. basketball) simplify the model, while sports with overtime or shoot-outs (e.g. ice hockey) can reuse the extra time stage. Adapting the model to other sports is a natural direction for future work, and the repository is designed to host ratings for other sports as well.

The gain of the average of the Glicko and the Markov models suggests that the two approaches capture different aspects of team strength. A natural next step is to analyze in which matches each of them is more accurate (for example, by the history of the teams, the gap between their strengths, the time since a transfer window, or the stage of the season) and to use these findings to build a single model that combines the stable league-aware ratings with the short-term information of the goal model, rather than averaging the predictions of two separate models.

In the evaluation, the parameters are estimated once on the seasons before 2021 and are not updated during the five test seasons. In practice, they would be re-estimated regularly, for example before every season. An expanding-window evaluation, in which the parameters are re-estimated before every test season on all earlier seasons, would show the accuracy of the model as it is used in practice and how quickly the parameters become outdated.

The collected data contain the scores by period for about two thirds of the matches with extra time, but not the minutes of the goals. With the minutes, the scoring rates of the Markov model at every score could be identified much better. Collecting the minutes of the goals is planned for the next version of the dataset.

The international matches of some leagues are not calibrated as well as those of the others: the clubs of Japan win more often than the model predicts, and the clubs of Iran less often (Section~\ref{subsec:calibration}). Whether these deviations come from the data or from the model remains to be checked.

Despite these constraints, the proposed framework remains computationally efficient and relies solely on publicly available data. While alternative probabilistic score-distribution models yield only marginal performance differences, the modified Glicko framework consistently improves predictive accuracy and provides interpretable ratings that can be directly applied to downstream analytical tasks.

\clearpage
\appendix

\section{The Pipeline of the Rating Estimation}\label{app:pipeline}

\begin{figure}[!htbp]
\centering
\includegraphics[height=0.8\textheight,keepaspectratio]{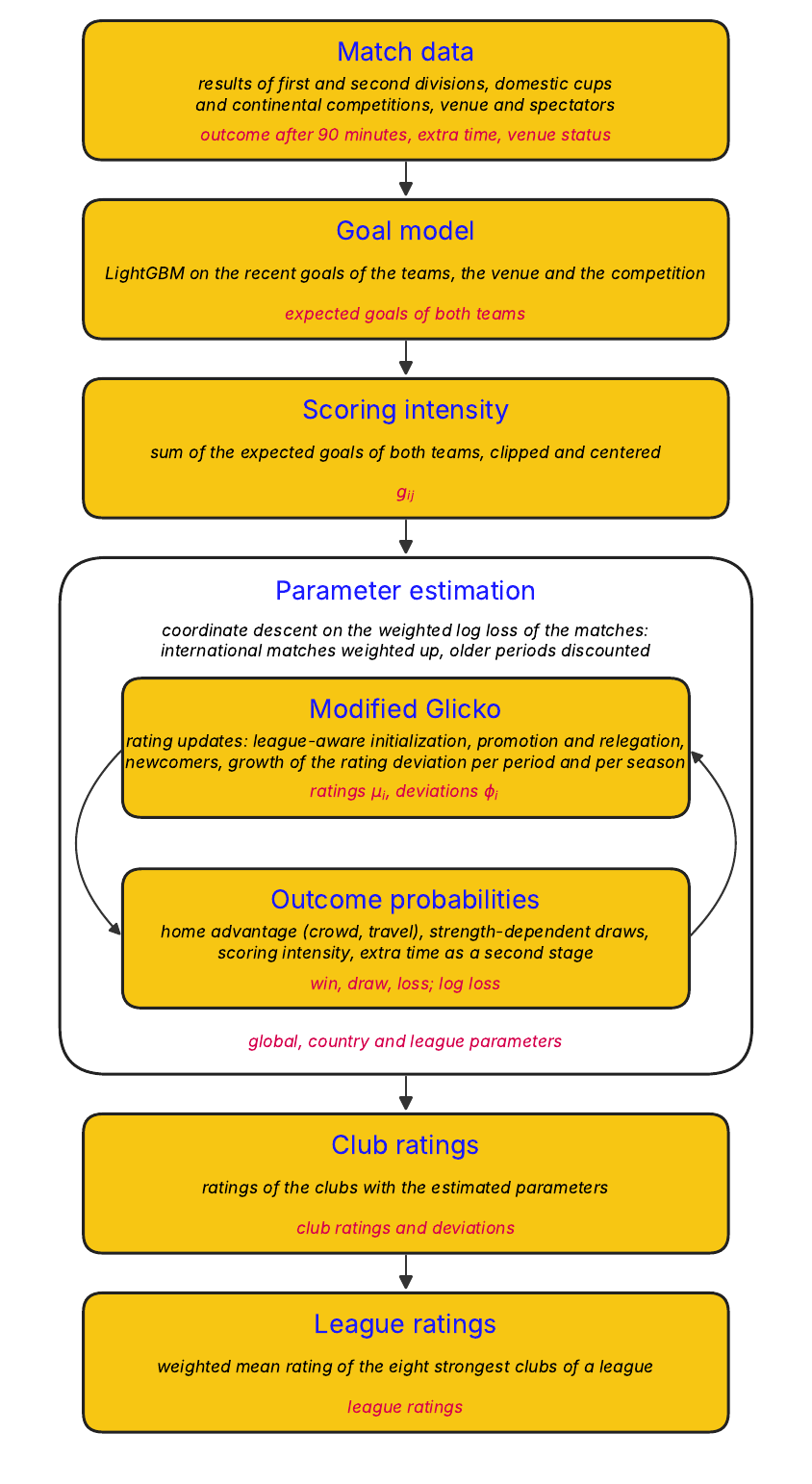}
\caption{The pipeline of the estimation of the club and league ratings}
\label{fig:pipeline_clubs}
\end{figure}

\begin{figure}[!htbp]
\centering
\includegraphics[height=0.7\textheight,keepaspectratio]{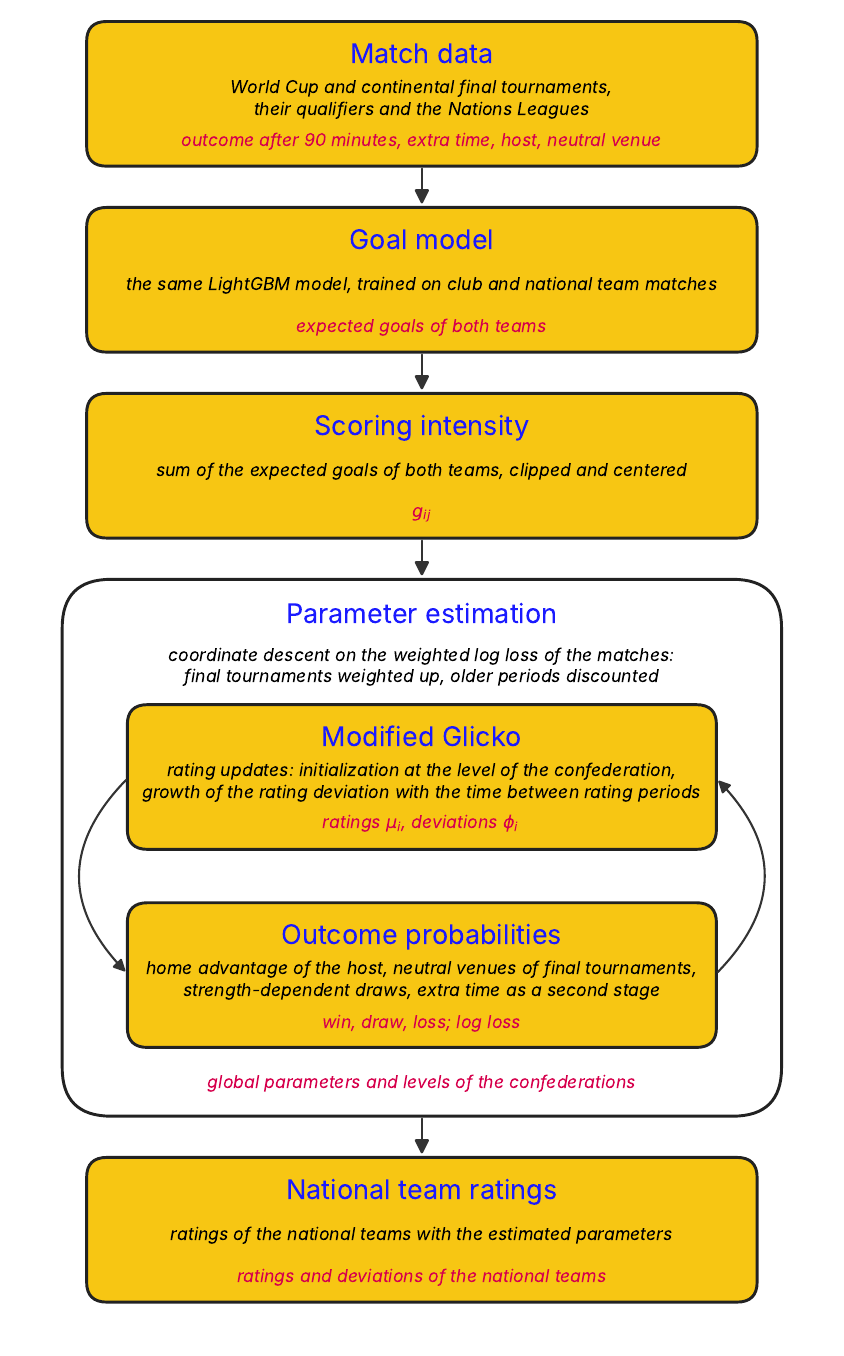}
\caption{The pipeline of the estimation of the national team ratings}
\label{fig:pipeline_national}
\end{figure}

\clearpage
\section{Ratings of Leagues, Clubs and National Teams}\label{app:ratings}
The ratings after the 2025/2026 season and the 2026 World Cup, computed with the parameters estimated on all seasons. The full tables are available in the repository.

{\sffamily\footnotesize\setlength{\tabcolsep}{3pt}\renewcommand{\arraystretch}{1.25}\setlength{\LTleft}{0pt}\setlength{\LTright}{0pt}\rowcolors{2}{Z}{white}
\begin{longtable}{rp{2.9cm}cccrlcl}
\caption[Ratings of the leagues]{Ratings of the first and second divisions after the 2025/2026 season: the weighted mean of the eight strongest clubs, Eq.~\eqref{eq:league_rating}. The stripe: the confederation (\stripe{1F5FA8}\,UEFA \stripe{2E8B3E}\,CONMEBOL \stripe{C0392B}\,AFC \stripe{D68910}\,CAF \stripe{7D3C98}\,CONCACAF \stripe{148F9A}\,OFC); SD: the standard deviation of the club ratings (darker: less even); n: the number of clubs; Int: the matches of the clubs of the league against foreign clubs in the last three seasons; the bar: from the median club to the strongest one, the dot: the league rating; the history of the league rating since 2016, relative to the world mean of every season; the strongest club and its world rank. Not shown: 6 leagues without matches against other leagues and without promotion or relegation in the last three seasons (Section~\ref{subsec:league_ratings}) and the Russian leagues}\label{tab:league_ratings}\\
\rowcolor{M}\textcolor{white}{\bfseries \#} & \textcolor{white}{\bfseries League} & \textcolor{white}{\bfseries Rating} & \textcolor{white}{\bfseries SD} & \textcolor{white}{\bfseries n} & \textcolor{white}{\bfseries Int} & \axisw{2.6} & \textcolor{white}{\bfseries 2016--2025} & \textcolor{white}{\bfseries Best club, rank}\\\endfirsthead
\rowcolor{M}\textcolor{white}{\bfseries \#} & \textcolor{white}{\bfseries League} & \textcolor{white}{\bfseries Rating} & \textcolor{white}{\bfseries SD} & \textcolor{white}{\bfseries n} & \textcolor{white}{\bfseries Int} & \axisw{2.6} & \textcolor{white}{\bfseries 2016--2025} & \textcolor{white}{\bfseries Best club, rank}\\\endhead
\arrayrulecolor{M}\specialrule{1pt}{0pt}{0pt}\endfoot
1 & \stripe{1F5FA8}\hspace{4pt}England I & \chip{F08A39}{2441} & \chip{469A9C}{118} & \textbf{20} & 214 & \rng{1.881}{2.426}{2.197}{F08A39}{M}{2.7} & \sparknd[1.65]{(0.000,0.139) -- (0.178,0.178) -- (0.356,0.242) -- (0.533,0.227) -- (0.711,0.239) -- (0.889,0.275) -- (1.067,0.273) -- (1.244,0.299) -- (1.422,0.286) -- (1.600,0.301)}{}{F08A39} & {\scriptsize\textcolor{muted}{Arsenal (2)}}\\
2 & \stripe{1F5FA8}\hspace{4pt}Spain I & \chip{E67A48}{2361} & \chip{36898B}{133} & \textbf{20} & 191 & \rng{1.645}{2.385}{2.043}{E67A48}{M}{2.7} & \sparknd[1.65]{(0.000,0.214) -- (0.178,0.198) -- (0.356,0.173) -- (0.533,0.175) -- (0.711,0.212) -- (0.889,0.201) -- (1.067,0.200) -- (1.244,0.267) -- (1.422,0.265) -- (1.600,0.238)}{}{E67A48} & {\scriptsize\textcolor{muted}{Barcelona (3)}}\\
3 & \stripe{1F5FA8}\hspace{4pt}Germany I & \chip{E4774A}{2345} & \chip{1E7073}{153} & \textbf{18} & 173 & \rng{1.585}{2.475}{2.013}{E4774A}{M}{2.7} & \sparknd[1.65]{(0.000,0.141) -- (0.178,0.128) -- (0.356,0.175) -- (0.533,0.231) -- (0.711,0.249) -- (0.889,0.214) -- (1.067,0.252) -- (1.244,0.271) -- (1.422,0.270) -- (1.600,0.312)}{}{E4774A} & {\scriptsize\textcolor{muted}{Bayern Munich (1)}}\\
4 & \stripe{1F5FA8}\hspace{4pt}Italy I & \chip{E2754D}{2332} & \chip{1E7073}{153} & \textbf{20} & 176 & \rng{1.596}{2.229}{1.988}{E2754D}{M}{2.7} & \sparknd[1.65]{(0.000,0.141) -- (0.178,0.207) -- (0.356,0.183) -- (0.533,0.199) -- (0.711,0.241) -- (0.889,0.235) -- (1.067,0.230) -- (1.244,0.257) -- (1.422,0.299) -- (1.600,0.276)}{}{E2754D} & {\scriptsize\textcolor{muted}{Inter (7)}}\\
5 & \stripe{1F5FA8}\hspace{4pt}France I & \chip{DC6B56}{2283} & \chip{2D8083}{141} & \textbf{18} & 155 & \rng{1.549}{2.276}{1.893}{DC6B56}{M}{2.7} & \sparknd[1.65]{(0.000,0.158) -- (0.178,0.170) -- (0.356,0.122) -- (0.533,0.147) -- (0.711,0.194) -- (0.889,0.221) -- (1.067,0.238) -- (1.244,0.257) -- (1.422,0.308) -- (1.600,0.318)}{}{DC6B56} & {\scriptsize\textcolor{muted}{PSG (6)}}\\
\end{longtable}}

\clearpage
\begin{table}[p]\sffamily\footnotesize\setlength{\tabcolsep}{2.3pt}\renewcommand{\arraystretch}{0.85}
\caption{The 100 strongest clubs after the 2025/2026 season. RD: the rating deviation (darker: less certain); the history since 2016, relative to the world mean of every season}\label{tab:club_ratings}
\rowcolors{2}{Z}{white}\begin{tabular}{rlccl@{\hspace{10pt}}rlccl}
\rowcolor{M}\textcolor{white}{\bfseries \#} & \textcolor{white}{\bfseries Club} & \textcolor{white}{\bfseries Rating} & \textcolor{white}{\bfseries RD} & \textcolor{white}{\bfseries 2016--2025} & \textcolor{white}{\bfseries \#} & \textcolor{white}{\bfseries Club} & \textcolor{white}{\bfseries Rating} & \textcolor{white}{\bfseries RD} & \textcolor{white}{\bfseries 2016--2025}\rule[-0.9ex]{0pt}{3ex}\\
1 & \stripe{1F5FA8}\hspace{4pt}Bayern Munich {\scriptsize\textcolor{muted}{Germany I}} & \chip{EE863D}{2585} & \chip[1.9em]{338689}{54} & \sparknd[1.35]{(0.000,0.166) -- (0.144,0.156) -- (0.289,0.133) -- (0.433,0.180) -- (0.578,0.211) -- (0.722,0.152) -- (0.867,0.136) -- (1.011,0.061) -- (1.156,0.107) -- (1.300,0.279)}{}{EE863D} & 51 & \stripe{1F5FA8}\hspace{4pt}Freiburg {\scriptsize\textcolor{muted}{Germany I}} & \chip{743D9E}{2215} & \chip[1.9em]{74BABC}{51} & \sparknd[1.35]{(0.000,0.108) -- (0.144,0.041) -- (0.289,0.020) -- (0.433,0.087) -- (0.578,0.124) -- (0.722,0.257) -- (0.867,0.320) -- (1.011,0.230) -- (1.156,0.284) -- (1.300,0.286)}{}{743D9E}\\
2 & \stripe{1F5FA8}\hspace{4pt}Arsenal {\scriptsize\textcolor{muted}{England I}} & \chip{E77C46}{2560} & \chip[1.9em]{8AC7C9}{51} & \sparknd[1.35]{(0.000,0.116) -- (0.144,0.041) -- (0.289,0.060) -- (0.433,0.020) -- (0.578,0.058) -- (0.722,0.072) -- (0.867,0.152) -- (1.011,0.227) -- (1.156,0.233) -- (1.300,0.320)}{}{E77C46} & 52 & \stripe{1F5FA8}\hspace{4pt}Real Sociedad {\scriptsize\textcolor{muted}{Spain I}} & \chip{733C9D}{2210} & \chip[1.9em]{57A8AA}{52} & \sparknd[1.35]{(0.000,0.101) -- (0.144,0.045) -- (0.289,0.072) -- (0.433,0.146) -- (0.578,0.191) -- (0.722,0.213) -- (0.867,0.295) -- (1.011,0.287) -- (1.156,0.174) -- (1.300,0.164)}{}{733C9D}\\
\end{tabular}\end{table}
\clearpage

{\sffamily\footnotesize\setlength{\tabcolsep}{4.1pt}\renewcommand{\arraystretch}{1.12}\captionof{table}{Ratings of the 216 national teams after the 2026 World Cup. The stripe: the confederation; RD: the rating deviation (darker: less certain); the history since 2010 at the end of every year, relative to the mean of all teams}\label{tab:national_ratings}
\noindent\rowcolors{2}{Z}{white}\begin{tabular}{@{}rlccl@{\hspace{8pt}}rlccl@{}}
\rowcolor{M}\textcolor{white}{\bfseries \#} & \textcolor{white}{\bfseries Team} & \textcolor{white}{\bfseries Rating} & \textcolor{white}{\bfseries RD} & \textcolor{white}{\bfseries 2010--2026} & \textcolor{white}{\bfseries \#} & \textcolor{white}{\bfseries Team} & \textcolor{white}{\bfseries Rating} & \textcolor{white}{\bfseries RD} & \textcolor{white}{\bfseries 2010--2026}\\
1 & \stripe{1F5FA8}\hspace{4pt}Spain & \chip{F28C38}{2415} & \chip[1.9em]{A8DADC}{72} & \sparknd[2.15]{(0.000,0.318) -- (0.131,0.339) -- (0.263,0.350) -- (0.394,0.346) -- (0.525,0.251) -- (0.656,0.278) -- (0.788,0.230) -- (0.919,0.256) -- (1.050,0.172) -- (1.181,0.164) -- (1.312,0.145) -- (1.444,0.096) -- (1.575,0.050) -- (1.706,0.058) -- (1.838,0.149) -- (1.969,0.138) -- (2.100,0.183)}{}{F28C38} & 37 & \stripe{1F5FA8}\hspace{4pt}Poland & \chip{C24579}{1902} & \chip[1.9em]{7FC0C2}{77} & \sparknd[2.15]{(0.000,0.205) -- (0.131,0.205) -- (0.263,0.216) -- (0.394,0.121) -- (0.525,0.215) -- (0.656,0.228) -- (0.788,0.336) -- (0.919,0.350) -- (1.050,0.284) -- (1.181,0.304) -- (1.312,0.280) -- (1.444,0.210) -- (1.575,0.211) -- (1.706,0.112) -- (1.838,0.053) -- (1.969,0.063) -- (2.100,0.050)}{}{C24579}\\
\end{tabular}\clearpage
\noindent\rowcolors{2}{Z}{white}\begin{tabular}{@{}rlccl@{\hspace{8pt}}rlccl@{}}
\rowcolor{M}\textcolor{white}{\bfseries \#} & \textcolor{white}{\bfseries Team} & \textcolor{white}{\bfseries Rating} & \textcolor{white}{\bfseries RD} & \textcolor{white}{\bfseries 2010--2026} & \textcolor{white}{\bfseries \#} & \textcolor{white}{\bfseries Team} & \textcolor{white}{\bfseries Rating} & \textcolor{white}{\bfseries RD} & \textcolor{white}{\bfseries 2010--2026}\\
73 & \stripe{1F5FA8}\hspace{4pt}Iceland & \chip{A44289}{1693} & \chip[1.9em]{70B7B9}{79} & \sparknd[2.15]{(0.000,0.107) -- (0.131,0.101) -- (0.263,0.116) -- (0.394,0.165) -- (0.525,0.223) -- (0.656,0.244) -- (0.788,0.312) -- (0.919,0.350) -- (1.050,0.281) -- (1.181,0.278) -- (1.312,0.219) -- (1.444,0.143) -- (1.575,0.131) -- (1.706,0.070) -- (1.838,0.077) -- (1.969,0.050) -- (2.100,0.050)}{}{A44289} & 109 & \stripe{D68910}\hspace{4pt}Rwanda & \chip{8E4095}{1534} & \chip[1.9em]{297C7F}{86} & \sparknd[2.15]{(0.000,0.318) -- (0.131,0.350) -- (0.263,0.317) -- (0.394,0.240) -- (0.525,0.285) -- (0.656,0.241) -- (0.788,0.195) -- (0.919,0.158) -- (1.050,0.125) -- (1.181,0.089) -- (1.312,0.106) -- (1.444,0.056) -- (1.575,0.050) -- (1.706,0.099) -- (1.838,0.135) -- (1.969,0.088) -- (2.100,0.088)}{}{8E4095}\\
\end{tabular}\clearpage
\noindent\rowcolors{2}{Z}{white}\begin{tabular}{@{}rlccl@{\hspace{8pt}}rlccl@{}}
\rowcolor{M}\textcolor{white}{\bfseries \#} & \textcolor{white}{\bfseries Team} & \textcolor{white}{\bfseries Rating} & \textcolor{white}{\bfseries RD} & \textcolor{white}{\bfseries 2010--2026} & \textcolor{white}{\bfseries \#} & \textcolor{white}{\bfseries Team} & \textcolor{white}{\bfseries Rating} & \textcolor{white}{\bfseries RD} & \textcolor{white}{\bfseries 2010--2026}\\
145 & \stripe{7D3C98}\hspace{4pt}Guyana & \chip{723C9D}{1331} & \chip[1.9em]{3B8E90}{83} & \sparknd[2.15]{(0.000,0.050) -- (0.131,0.350) -- (0.263,0.275) -- (0.394,0.275) -- (0.525,0.275) -- (0.656,0.246) -- (0.788,0.246) -- (0.919,0.246) -- (1.050,0.212) -- (1.181,0.252) -- (1.312,0.252) -- (1.444,0.136) -- (1.575,0.159) -- (1.706,0.214) -- (1.838,0.170) -- (1.969,0.185) -- (2.100,0.200)}{}{723C9D} & 181 & \stripe{7D3C98}\hspace{4pt}Saint Martin & \chip{513293}{1064} & \chip[1.9em]{155961}{94} & \sparknd[2.15]{(0.000,0.350) -- (0.131,0.350) -- (0.263,0.350) -- (0.394,0.350) -- (0.525,0.350) -- (0.656,0.350) -- (0.788,0.350) -- (0.919,0.350) -- (1.050,0.149) -- (1.181,0.081) -- (1.312,0.081) -- (1.444,0.081) -- (1.575,0.050) -- (1.706,0.092) -- (1.838,0.092) -- (1.969,0.085) -- (2.100,0.109)}{}{513293}\\
\end{tabular}\clearpage}
\clearpage

\bmhead{Acknowledgements}

Andrei Shelopugin is thankful to Nikita Kozodoi and Viktoria Lokteva for their contributions to the earlier stages of this research, and to Nikita Vasyukhin for consultations on soccer.

\bibliographystyle{sn-mathphys-num}
\bibliography{Ratings}


\begin{thebibliography}{20}
\ifx \bisbn   \undefined \def \bisbn  #1{ISBN #1}\fi
\ifx \binits  \undefined \def \binits#1{#1}\fi
\ifx \bauthor  \undefined \def \bauthor#1{#1}\fi
\ifx \batitle  \undefined \def \batitle#1{#1}\fi
\ifx \bjtitle  \undefined \def \bjtitle#1{#1}\fi
\ifx \bvolume  \undefined \def \bvolume#1{\textbf{#1}}\fi
\ifx \byear  \undefined \def \byear#1{#1}\fi
\ifx \bissue  \undefined \def \bissue#1{#1}\fi
\ifx \bfpage  \undefined \def \bfpage#1{#1}\fi
\ifx \blpage  \undefined \def \blpage #1{#1}\fi
\ifx \burl  \undefined \def \burl#1{\textsf{#1}}\fi
\ifx \doiurl  \undefined \def \doiurl#1{\url{https://doi.org/#1}}\fi
\ifx \betal  \undefined \def \betal{\textit{et al.}}\fi
\ifx \binstitute  \undefined \def \binstitute#1{#1}\fi
\ifx \binstitutionaled  \undefined \def \binstitutionaled#1{#1}\fi
\ifx \bctitle  \undefined \def \bctitle#1{#1}\fi
\ifx \beditor  \undefined \def \beditor#1{#1}\fi
\ifx \bpublisher  \undefined \def \bpublisher#1{#1}\fi
\ifx \bbtitle  \undefined \def \bbtitle#1{#1}\fi
\ifx \bedition  \undefined \def \bedition#1{#1}\fi
\ifx \bseriesno  \undefined \def \bseriesno#1{#1}\fi
\ifx \blocation  \undefined \def \blocation#1{#1}\fi
\ifx \bsertitle  \undefined \def \bsertitle#1{#1}\fi
\ifx \bsnm \undefined \def \bsnm#1{#1}\fi
\ifx \bsuffix \undefined \def \bsuffix#1{#1}\fi
\ifx \bparticle \undefined \def \bparticle#1{#1}\fi
\ifx \barticle \undefined \def \barticle#1{#1}\fi
\bibcommenthead
\ifx \bconfdate \undefined \def \bconfdate #1{#1}\fi
\ifx \botherref \undefined \def \botherref #1{#1}\fi
\ifx \url \undefined \def \url#1{\textsf{#1}}\fi
\ifx \bchapter \undefined \def \bchapter#1{#1}\fi
\ifx \bbook \undefined \def \bbook#1{#1}\fi
\ifx \bcomment \undefined \def \bcomment#1{#1}\fi
\ifx \oauthor \undefined \def \oauthor#1{#1}\fi
\ifx \citeauthoryear \undefined \def \citeauthoryear#1{#1}\fi
\ifx \endbibitem  \undefined \def \endbibitem {}\fi
\ifx \bconflocation  \undefined \def \bconflocation#1{#1}\fi
\ifx \arxivurl  \undefined \def \arxivurl#1{\textsf{#1}}\fi
\csname PreBibitemsHook\endcsname

\bibitem[\protect\citeauthoryear{Bergkamp et~al.}{2022}]{bergkamp2022soccer}
\begin{barticle}
\bauthor{\bsnm{Bergkamp}, \binits{T.L.}},
\bauthor{\bsnm{Frencken}, \binits{W.G.}},
\bauthor{\bsnm{Niessen}, \binits{A.S.M.}},
\bauthor{\bsnm{Meijer}, \binits{R.R.}},
\bauthor{\bsnm{Hartigh}, \binits{R.J.}}:
\batitle{How soccer scouts identify talented players}.
\bjtitle{European Journal of Sport Science}
\bvolume{22}(\bissue{7}),
\bfpage{994}--\blpage{1004}
(\byear{2022})
\end{barticle}
\endbibitem

\bibitem[\protect\citeauthoryear{Dixon and Coles}{1997}]{dixon1997modelling}
\begin{barticle}
\bauthor{\bsnm{Dixon}, \binits{M.J.}},
\bauthor{\bsnm{Coles}, \binits{S.G.}}:
\batitle{Modelling association football scores and inefficiencies in the
  football betting market}.
\bjtitle{Journal of the Royal Statistical Society: Series C (Applied
  Statistics)}
\bvolume{46}(\bissue{2}),
\bfpage{265}--\blpage{280}
(\byear{1997})
\end{barticle}
\endbibitem

\bibitem[\protect\citeauthoryear{Elo}{1967}]{elo1967proposed}
\begin{barticle}
\bauthor{\bsnm{Elo}, \binits{A.E.}}:
\batitle{The proposed uscf rating system, its development, theory, and
  applications}.
\bjtitle{Chess Life}
\bvolume{22}(\bissue{8}),
\bfpage{242}--\blpage{247}
(\byear{1967})
\end{barticle}
\endbibitem

\bibitem[\protect\citeauthoryear{Glickman}{2026}]{glickman2026paired}
\begin{barticle}
\bauthor{\bsnm{Glickman}, \binits{M.E.}}:
\batitle{Paired comparison models with strength-dependent ties and order
  effects}.
\bjtitle{Statistical Modelling}
\bvolume{26}(\bissue{5}),
\bfpage{476}--\blpage{499}
(\byear{2026})
\end{barticle}
\endbibitem

\bibitem[\protect\citeauthoryear{Grinsztajn et~al.}{2022}]{grinsztajn2022tree}
\begin{barticle}
\bauthor{\bsnm{Grinsztajn}, \binits{L.}},
\bauthor{\bsnm{Oyallon}, \binits{E.}},
\bauthor{\bsnm{Varoquaux}, \binits{G.}}:
\batitle{Why do tree-based models still outperform deep learning on typical
  tabular data?}
\bjtitle{Advances in neural information processing systems}
\bvolume{35},
\bfpage{507}--\blpage{520}
(\byear{2022})
\end{barticle}
\endbibitem

\bibitem[\protect\citeauthoryear{Ke et~al.}{2017}]{ke2017lightgbm}
\begin{botherref}
\oauthor{\bsnm{Ke}, \binits{G.}},
\oauthor{\bsnm{Meng}, \binits{Q.}},
\oauthor{\bsnm{Finley}, \binits{T.}},
\oauthor{\bsnm{Wang}, \binits{T.}},
\oauthor{\bsnm{Chen}, \binits{W.}},
\oauthor{\bsnm{Ma}, \binits{W.}},
\oauthor{\bsnm{Ye}, \binits{Q.}},
\oauthor{\bsnm{Liu}, \binits{T.-Y.}}:
Lightgbm: A highly efficient gradient boosting decision tree.
Advances in neural information processing systems
\textbf{30}
(2017)
\end{botherref}
\endbibitem

\bibitem[\protect\citeauthoryear{Prokhorenkova
  et~al.}{2018}]{prokhorenkova2018catboost}
\begin{botherref}
\oauthor{\bsnm{Prokhorenkova}, \binits{L.}},
\oauthor{\bsnm{Gusev}, \binits{G.}},
\oauthor{\bsnm{Vorobev}, \binits{A.}},
\oauthor{\bsnm{Dorogush}, \binits{A.V.}},
\oauthor{\bsnm{Gulin}, \binits{A.}}:
Catboost: unbiased boosting with categorical features.
Advances in neural information processing systems
\textbf{31}
(2018)
\end{botherref}
\endbibitem

\bibitem[\protect\citeauthoryear{Dixon and Robinson}{1998}]{dixon1998birth}
\begin{barticle}
\bauthor{\bsnm{Dixon}, \binits{M.}},
\bauthor{\bsnm{Robinson}, \binits{M.}}:
\batitle{A birth process model for association football matches}.
\bjtitle{Journal of the Royal Statistical Society: Series D (The Statistician)}
\bvolume{47}(\bissue{3}),
\bfpage{523}--\blpage{538}
(\byear{1998})
\end{barticle}
\endbibitem

\bibitem[\protect\citeauthoryear{Glickman}{2012}]{glickman2012example}
\begin{botherref}
\oauthor{\bsnm{Glickman}, \binits{M.E.}}:
Example of the glicko-2 system.
Boston University
\textbf{28}
(2012)
\end{botherref}
\endbibitem

\bibitem[\protect\citeauthoryear{Herbrich et~al.}{2006}]{herbrich2006trueskill}
\begin{botherref}
\oauthor{\bsnm{Herbrich}, \binits{R.}},
\oauthor{\bsnm{Minka}, \binits{T.}},
\oauthor{\bsnm{Graepel}, \binits{T.}}:
Trueskill™: a bayesian skill rating system.
Advances in neural information processing systems
\textbf{19}
(2006)
\end{botherref}
\endbibitem

\bibitem[\protect\citeauthoryear{Shelopugin and
  Sirotkin}{2023}]{shelopugin2023ratings}
\begin{botherref}
\oauthor{\bsnm{Shelopugin}, \binits{A.}},
\oauthor{\bsnm{Sirotkin}, \binits{A.}}:
Ratings of european and south american football leagues based on glicko-2 with
  modifications.
arXiv preprint arXiv:2310.11459
(2023)
\end{botherref}
\endbibitem

\bibitem[\protect\citeauthoryear{Shelopugin and
  Sirotkin}{2026}]{shelopugin2026constructing}
\begin{bchapter}
\bauthor{\bsnm{Shelopugin}, \binits{A.}},
\bauthor{\bsnm{Sirotkin}, \binits{A.}}:
\bctitle{Constructing football club ratings with league strength awareness
  using a modified glicko model}.
In: \bbtitle{International Sports Analytics Conference and Exhibition},
pp. \bfpage{38}--\blpage{61}
(\byear{2026}).
\bcomment{Springer}
\end{bchapter}
\endbibitem

\bibitem[\protect\citeauthoryear{Pollard}{1986}]{pollard1986home}
\begin{barticle}
\bauthor{\bsnm{Pollard}, \binits{R.}}:
\batitle{Home advantage in soccer: a retrospective analysis}.
\bjtitle{Journal of Sports Sciences}
\bvolume{4}(\bissue{3}),
\bfpage{237}--\blpage{248}
(\byear{1986})
\end{barticle}
\endbibitem

\bibitem[\protect\citeauthoryear{Nevill et~al.}{2002}]{nevill2002influence}
\begin{barticle}
\bauthor{\bsnm{Nevill}, \binits{A.M.}},
\bauthor{\bsnm{Balmer}, \binits{N.J.}},
\bauthor{\bsnm{Williams}, \binits{A.M.}}:
\batitle{The influence of crowd noise and experience upon refereeing decisions
  in football}.
\bjtitle{Psychology of Sport and Exercise}
\bvolume{3}(\bissue{4}),
\bfpage{261}--\blpage{272}
(\byear{2002})
\end{barticle}
\endbibitem

\bibitem[\protect\citeauthoryear{Pettersson-Lidbom and
  Priks}{2010}]{pettersson2010behavior}
\begin{barticle}
\bauthor{\bsnm{Pettersson-Lidbom}, \binits{P.}},
\bauthor{\bsnm{Priks}, \binits{M.}}:
\batitle{Behavior under social pressure: Empty italian stadiums and referee
  bias}.
\bjtitle{Economics Letters}
\bvolume{108}(\bissue{2}),
\bfpage{212}--\blpage{214}
(\byear{2010})
\end{barticle}
\endbibitem

\bibitem[\protect\citeauthoryear{Constantinou and
  Fenton}{2012}]{constantinou2012solving}
\begin{botherref}
\oauthor{\bsnm{Constantinou}, \binits{A.C.}},
\oauthor{\bsnm{Fenton}, \binits{N.E.}}:
Solving the problem of inadequate scoring rules for assessing probabilistic
  football forecast models.
Journal of Quantitative Analysis in Sports
\textbf{8}(1)
(2012)
\end{botherref}
\endbibitem

\bibitem[\protect\citeauthoryear{Davidson}{1970}]{davidson1970extending}
\begin{barticle}
\bauthor{\bsnm{Davidson}, \binits{R.R.}}:
\batitle{On extending the bradley-terry model to accommodate ties in paired
  comparison experiments}.
\bjtitle{Journal of the American Statistical Association}
\bvolume{65}(\bissue{329}),
\bfpage{317}--\blpage{328}
(\byear{1970})
\end{barticle}
\endbibitem

\bibitem[\protect\citeauthoryear{Dinsdale and
  Gallagher}{2022}]{dinsdale2022transfer}
\begin{botherref}
\oauthor{\bsnm{Dinsdale}, \binits{D.}},
\oauthor{\bsnm{Gallagher}, \binits{J.}}:
Transfer portal: Accurately forecasting the impact of a player transfer in
  soccer.
arXiv preprint arXiv:2201.11533
(2022)
\end{botherref}
\endbibitem

\bibitem[\protect\citeauthoryear{Shelopugin}{2026}]{shelopugin2026expected}
\begin{bchapter}
\bauthor{\bsnm{Shelopugin}, \binits{A.}}:
\bctitle{Expected possession value of control and duel actions for soccer
  player skills estimation}.
In: \bbtitle{International Sports Analytics Conference and Exhibition},
pp. \bfpage{292}--\blpage{314}
(\byear{2026}).
\bcomment{Springer}
\end{bchapter}
\endbibitem

\bibitem[\protect\citeauthoryear{Pollard and Reep}{1997}]{pollard1997measuring}
\begin{barticle}
\bauthor{\bsnm{Pollard}, \binits{R.}},
\bauthor{\bsnm{Reep}, \binits{C.}}:
\batitle{Measuring the effectiveness of playing strategies at soccer}.
\bjtitle{Journal of the Royal Statistical Society Series D: The Statistician}
\bvolume{46}(\bissue{4}),
\bfpage{541}--\blpage{550}
(\byear{1997})
\end{barticle}
\endbibitem

\end{thebibliography}

\end{document}